\documentclass[runningheads]{llncs}

\usepackage[review,year=2026,ID=2090]{eccv}

\usepackage{eccvabbrv}

\usepackage{graphicx}
\usepackage{booktabs}

\usepackage[accsupp]{axessibility}  

\usepackage{graphicx,verbatim}
\usepackage{adjustbox}
\usepackage{multirow}
\usepackage{bm}
\usepackage[table, dvipsnames, svgnames, x11names, HTML]{xcolor}
\usepackage{tcolorbox}
\usepackage{graphicx,verbatim, xspace, amsmath, amsfonts, algorithm, algorithmicx, algpseudocode}
\usepackage{array, booktabs}
\usepackage{tikz}
\usetikzlibrary{positioning}
\usepackage{graphicx}

\usepackage{orcidlink}
\usepackage{wrapfig}
\usepackage{makecell}
\usepackage{caption}
\usepackage{adjustbox}
\usepackage{booktabs}
\usepackage{graphicx} 
\newcommand{\ours}{MedSyn2}
\definecolor{cvprblue}{rgb}{0.21,0.49,0.74}
\begin{document}

\title{MedSyn2: Flexible Control of 3D CT Generation via Text and Semantically-Defined Segmentation Prompts} 

\titlerunning{Abbreviated paper title}

\author{
Weicheng Dai\inst{1}\and
Chenyu Wang\inst{1}\and
Binxu Li\inst{2}\and
Shantanu Ghosh\inst{1}\and
Afrooz Zandifar\inst{3}\and 
Christina LeBedis\inst{1,4}\and 
Kayhan Batmanghelich\inst{1}
}

\authorrunning{W. Dai et al.}

\institute{
Boston University School of Engineering, Boston, MA, USA\\
\email{\{wd2119, chyuwang, shawn24, batman\}@bu.edu}
\and
Stanford University\\
\email{andy0207@stanford.edu}
\and
University of Pittsburgh Medical Center \and Boston University School of Medicine
}



\maketitle

\begin{abstract}
Generative models for volumetric medical images have found many applications in medical imaging, ranging from data augmentation to serving as priors for inverse problems. 
For these applications, generating high-resolution 3D images with strong controllability is essential but remains highly challenging. 
Existing approaches typically control generation either through radiology reports used as text prompts or through full image segmentation. 
While text-based prompting is flexible, it provides limited spatial control over the location, shape, and boundary of abnormalities. 
In contrast, segmentation-based methods receive precise spatial guidance but are restrictive in requiring full-organ annotations.
In this work, we propose \ours, a flexible multimodal framework for controllable volumetric image generation that supports input from radiology reports and segmentation prompts (both optional). 
Our approach allows users to provide segmentation of a specific anatomy or abnormality without requiring full-organ annotations. 
The semantic meaning of the segmentation mask is specified through an accompanying text description, resulting in a highly flexible and scalable conditioning mechanism. 
We develop a memory-efficient architecture based on a modified diffusion transformer that jointly processes image and segmentation tokens. 
The model further incorporates gated attention to effectively attend to long radiology reports.
Experiments demonstrate that our method achieves state-of-the-art perceptual and semantic scores (\eg, $\sim 24\%$ relative improvement in mean FID), generates high-resolution anatomically consistent CT volumes, and improves data efficiency when used for data augmentation. 
Radiologists' evaluation further confirms strong alignment between generated and real medical images. 
We also explore the ability of the model to generalize towards concepts beyond training data.
\textit{Code will be released upon decision.}
\end{abstract}

\section{Introduction}

Generative models play an important role in medical imaging. They enable privacy-preserving data augmentation~\cite{generatect}, serve as building blocks for model explanation~\cite{kumar2025prism}, provide strong priors for solving inverse problems~\cite{chung2023solving}, and support the training of radiologist residents~\cite{katelyn}. 
For these applications, controlling the generation process is essential, as the semantic properties of the synthetic image must be specified explicitly. 
Most existing approaches control generation using radiology reports as textual prompts~\cite{generatect, medsyn2024, molino2025texttoctgeneration3dlatent, text2CTGuo, amirrajab2025radiologyreportconditional3d, wang2025ctflowvideoinspiredlatentflow, Liu_2024_ACCV}. 
However, radiology reports provide limited control over spatial attributes, \eg they cannot precisely specify the shape of a nodule or the precise size of opacity. 
Such problems can be mitigated by segmentation masks, 
which clearly define boundaries of pathologies and anatomies.
Thus other methods use masks for synthesis~\cite{medddpm, zhao2025maisiv2accelerated3dhighresolution, maisiv1, Wang20243DMA, HanMedGen3D, Krishna_2025, zhuangGen, oliveras2025land, lungddpm}, yet they require full annotations on all organs which is expensive. 
This paper proposes a flexible and scalable multimodal framework that controls high-resolution volumetric CT synthesis using either radiology reports or segmentation masks, or a combination of both, where the user specifies semantic meaning of the segmentation. 

Currently, there are two main ways to control volumetric radiology image generation: using radiology reports as prompts~\cite{generatect, medsyn2024, molino2025texttoctgeneration3dlatent, text2CTGuo, amirrajab2025radiologyreportconditional3d, wang2025ctflowvideoinspiredlatentflow, Liu_2024_ACCV} or using segmentation masks as spatial inputs~\cite{medddpm, zhao2025maisiv2accelerated3dhighresolution, maisiv1, Wang20243DMA, HanMedGen3D, Krishna_2025, zhuangGen, oliveras2025land, lungddpm}. 
Radiology reports contain rich clinical information describing observed abnormalities. 
However, these descriptions are always written in approximate terms.
Text itself cannot precisely specify the location, shape, or boundary of an abnormality, although such details may carry important clinical value. 
Segmentation masks, on the other hand, alleviate the burden by defining clear spatial descriptions, helpful in defining sizes of abnormalities.
Methods that condition generation on segmentation masks address this limitation by providing explicit spatial information, where anatomical structures and lesion boundaries are defined precisely. 
However, current state-of-the-art approaches typically require full segmentation covering all organs and abnormalities in the volume (\eg, 117 classes in~\cite{zhao2025maisiv2accelerated3dhighresolution}). 
As a result, they are often limited to a small set of abnormality labels defined during training. 
Moreover, these approaches impose a heavy burden at inference time, as full-organ segmentation is required before generation. 
They also reduce the diversity of the generated images, since all organs and abnormalities must be predefined. 
Importantly, segmentation and radiology reports provide complementary information and should be ideally used together. 
Our goal therefore is to develop a method that combines the strengths of both modalities, enabling precise control while remaining flexible and scalable.

A desirable method should be \emph{flexible}, allowing users to provide segmentation of one specific abnormality or anatomy without requiring full-organ annotations. 
In addition, the semantic meaning of the segmentation prompt should be defined explicitly and flexibly, since a mask may represent different concepts such as a lung nodule, heart boundaries, or mask of pleural effusion, etc. 
This design increases the diversity of the generated images, as the model retains more degrees of freedom compared to approaches that rely on full segmentation where all boundaries are predefined. 
Our architecture builds on a modified diffusion transformer (DiT)~\cite{Peebles2022DiT}. 
We tokenize both the image and the segmentation mask, when available, with a fine-tuned video tokenization method. 
This design is memory efficient for high-resolution CT generation, as image and segmentation tokens are processed jointly without increasing the number of tokens, incurring negligible additional memory cost. 
To better handle long radiology reports, we incorporate gated attention~\cite{qiu2025gated} to ensure the model attends to all text tokens. 
The semantic meaning of the segmentation prompt is provided as optional text appended to the radiology report, enabling \emph{scalable} conditioning where new abnormalities can be introduced through fine-tuning or even in a zero-shot setting. 

This work makes three \textbf{contributions}. 
(1) We develop \ours, a framework where users can optionally provide a radiology report or a segmentation of an abnormality, or both to control image generation, resulting in a highly versatile generative model. 
(2) We propose an architecture that attends to long text prompts and segmentation while generating high-resolution radiology images on modest GPU resources. 
(3) Our method achieves state-of-the-art perceptual and semantic scores. 
We further show that the generated images exhibit high anatomical fidelity, improve data efficiency when used for data augmentation, and align well with real medical images according to evaluation by two radiologists.

\section{Related Work}

\noindent \textbf{Text-to-Image Lung CT Generation.}
Generating high-resolution 3D medical images is computationally expensive due to the large spatial dimensions of volumetric data. CTFlow~\cite{wang2025ctflowvideoinspiredlatentflow} mitigates this challenge by leveraging 2D backbones to generate slices and auto-regressively assemble them into volumes. 
While this design reduces training memory requirements, it often leads to long inference times and anatomical inconsistencies across orthogonal planes. 
GenerateCT~\cite{generatect} and MedSyn~\cite{medsyn2024} address memory limitations using multi-resolution generation strategies. 
However, these approaches synthesize images directly in the voxel domain, which becomes computationally expensive when extending to high-resolution volumetric images. 
More recent methods, including Text2CT~\cite{molino2025texttoctgeneration3dlatent}, T2CT~\cite{text2CTGuo}, and Report2CT~\cite{amirrajab2025radiologyreportconditional3d}, adopt latent diffusion models (LDM)~\cite{rombach2021highresolution}, which are more computationally scalable. 
Despite this improvement, these models rely solely on textual prompts and therefore inherit the limitations of text-only control, such as limited ability to specify the precise size, shape, or spatial extent of anatomical structures and abnormalities.
Such drawbacks can be mitigated via segmentation masks as they absorb more abundant spatial information.

\noindent \textbf{Segmentation-to-Image Lung CT Generation.}
Using semantic segmentation masks as prompts provides an alternative approach for controlling image generation. 
SemGen~\cite{zhuangGen}, mDDPM~\cite{Krishna_2025}, and MedGen3D~\cite{HanMedGen3D} generate CT images conditioned on segmentation masks using 2D backbones. 
However, similar to text-based methods built on 2D architectures, these approaches inherit limitations when generating consistent volumetric structures. 
LungDDPM~\cite{lungddpm} and LAND~\cite{oliveras2025land} instead adopt 3D backbones, but they focus on inpainting local abnormalities such as lung nodules in existing scans rather than synthesizing full CT volumes. 
Med-DDPM~\cite{medddpm} generates full volumes conditioned on paired images and semantic masks in the form of image-to-image style transfer. 
However, requiring a full volumetric mask as input limits the flexibility and applicability of the method for downstream tasks, since the model cannot generate complete images without training with corresponding masks. 
MAISIs~\cite{maisiv1, zhao2025maisiv2accelerated3dhighresolution} and 3D MedDiffusion~\cite{Wang20243DMA} attempt to alleviate this limitation by first training unconditional volumetric generators and then fine-tuning them using ControlNet~\cite{zhang2023adding} to accept segmentation masks as conditions. 
Nevertheless, these methods still require full segmentation masks covering all organs in the field of view. 
It reduces the diversity of the generated images and limits the model accepting segmentation labels only defined during fine-tuning, requiring retraining when new mask types are introduced. 
Importantly, none of these segmentation-based approaches incorporates radiology reports as prompts, even though textual reports provide complementary clinical information beyond spatial annotations. 
Recent joint-modality approaches (DGM~\cite{xing2024deep}, Trace~\cite{TRACE}) attempt to combine text and segmentation prompts, but similar to MAISIs they require full segmentation masks defining all organs, which limits flexibility in practical clinical scenarios.

\begin{figure}
    \centering
    \includegraphics[width=1\linewidth]{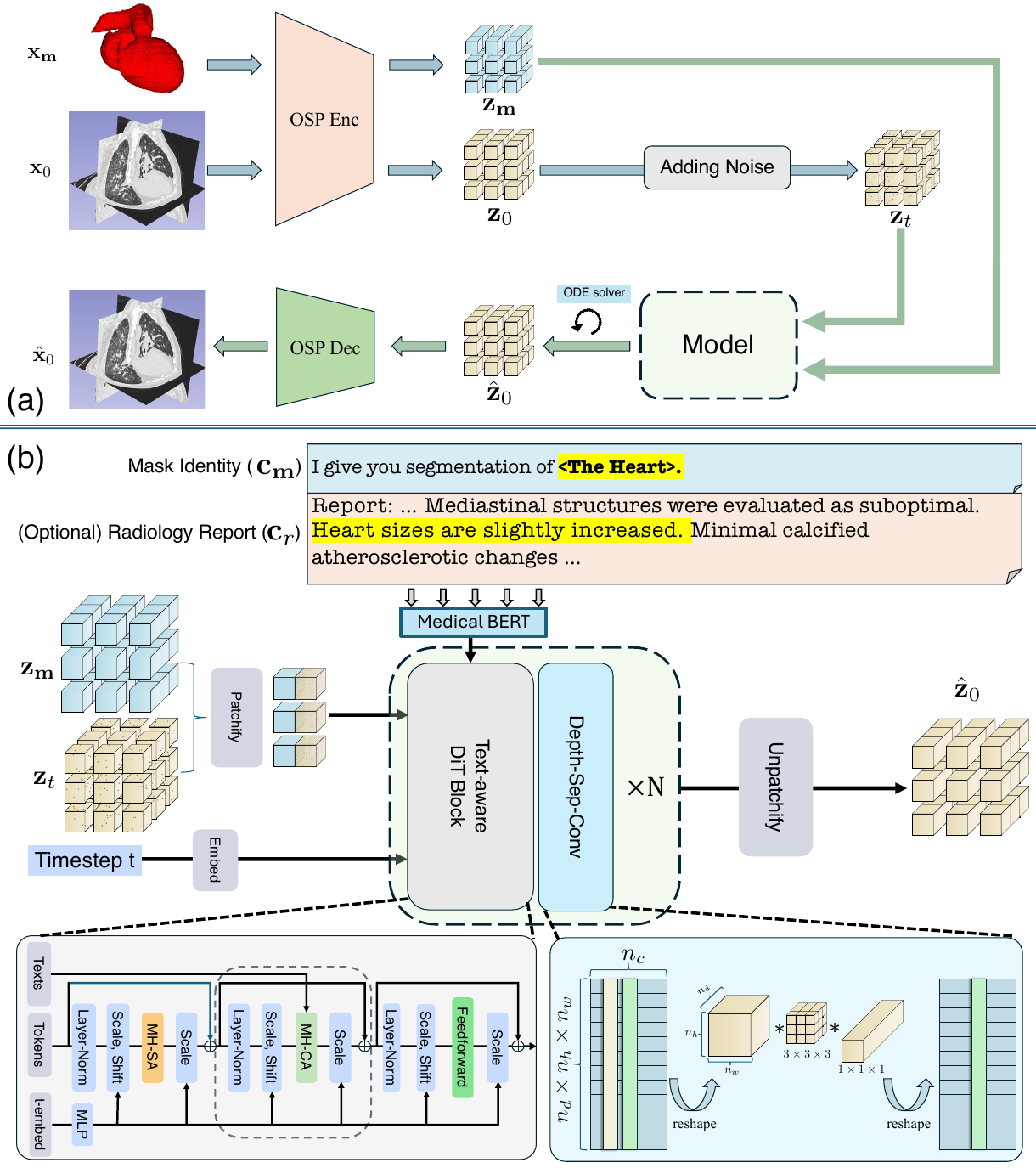}
    \caption{\textbf{Overview of \ours.}
    \textbf{(a.)} Our Encoder-Decoder utilizes OSP trained on CT-Rate images, while we inference on both $\mathbf{x}_0$ and $\mathbf{x}_\mathbf{m}$.
    \textbf{(b.)} We inject compound text embeddings ($\mathbf{c}_\mathbf{m}$, optional $\mathbf{c}_r$) into text-aware DiT Block with multi-head cross-attention.  
    The paired segmentation latent $\mathbf{z}_\mathbf{m}$ and noisy image latent $\mathbf{z}_t$ are patchified together in early stages to learn a clean image latent $\mathbf{z}_0$. 
    A light-weighted Depth separable convolution layer (Depth-Sep-Conv) is appended after each Block for maintaining inter-token feature consistency.
    }
    \label{fig:overview}
\end{figure}



\section{Method}

\noindent \textbf{Overview.}
Our goal is to develop a controllable framework for 3D CT synthesis that uses both long-form radiology reports and segmentation masks as prompts. 
The model should have the following capabilities: (1) A user should be able to control the image generation with either a radiology report, or a segmentation mask, or both. 
(2) The user can \emph{define} the semantic meaning of a segmentation mask with extra text appended to the radiology prompt, where both prompts are optional. 
(3) The model jointly attends to spatial and semantic conditions while preserving \emph{high-fidelity anatomical details} in \emph{high-resolution} volumes. 
We address these requirements via a rectified flow formulation~\cite{lipman2023flow} operating on latent tokens, where a diffusion transformer (DiT)~\cite{Peebles2022DiT} predicts the final clean latent state (i.e., $\mathbf{x}_0$ prediction) from noisy inputs. 
This setting introduces several challenges: 
(a) generating volumetric images is highly memory demanding, 
(b) radiology reports are long and standard architectures often focus on early tokens~\cite{qiu2025gated}, 
and (c) segmentation masks do not have fixed semantics and the user should be able to define the meaning of the mask prompt. 
To address these issues, we introduce a compound prompting strategy that explicitly encodes mask semantics as a textual description together with the radiology report.
We further address (a) through a unified tokenization scheme adapted from video tokenization (OSP~\cite{lin2024opensoraplanopensourcelarge}), which treats images and segmentations in the same latent space. 
We further address (b) by adopting gated attention~\cite{qiu2025gated} to improve reasoning over long and complex text prompts (Ablation in Supp.~\ref{supp:gated}). 
Finally, for (c) we modify the DiT architecture to jointly reason over segmentation and report conditioning while improving spatial consistency and anatomical fidelity in generated volumes. 
Fig.~\ref{fig:overview} shows the schematic of our model architecture.

\subsection{Data Preparation and Image Tokenization}
Our framework follows a latent diffusion formulation and therefore requires a tokenization strategy that preserves anatomical fidelity while efficiently handling volumetric data. 
Let $\text{Enc}(\cdot)$ denote the encoder used to tokenize volumetric inputs into latent representations. 
We denote a CT image volume by $\mathbf{x}$ and its corresponding segmentation mask by $\mathbf{m}$. 
Prior to tokenization, all CT volumes are affinely registered to a common template such that the initial and terminal slices approximately correspond to the same anatomical regions. 
This alignment reduces inter-subject spatial variance and stabilizes token learning. 
Given the aligned volumes, we interpret each 3D CT scan as videos alongside axial axis~\cite{text2CTGuo}, and adopt a video tokenization paradigm. 
Among several alternatives, we fine-tune the Open-Sora~\cite{opensora,opensora2} tokenizer on CT images to obtain compact yet high-fidelity latent representations (reconstruction analysis in Supp.~\ref{supp:tokenizer}).

The image tokens are defined as
$
\mathbf{z}_{0} = \text{Enc}(\mathbf{x}) \in 
\mathbb{R}^{(n_d \times n_h \times n_w) \times n_c},
$
where $n_d$, $n_h$, and $n_w$ denote the number of tokens along depth, height, and width, respectively, and $n_c$ is the latent channel dimension. 
Following prior work~\cite{wang2025lavin}, the tokenizer is fine-tuned using images only and is not trained on segmentation masks. 
During inference, however, the same encoder is applied to segmentation masks to obtain
$
\mathbf{z}_{\mathbf{m}} = \text{Enc}(\mathbf{m}) \in 
\mathbb{R}^{(n_d \times n_h \times n_w) \times n_c}.
$
This unified tokenization strategy compresses both images and segmentations within a shared latent space, allowing the diffusion model to process spatial and anatomical conditions in a consistent manner.



\subsection{Proposed Architecture}
\label{sec:model}

\noindent \textbf{Patchification and Compound Prompting.}
Given a noisy version of the tokenized image $\mathbf{z}_t$ at time $t$ and a clean segmentation representation $\mathbf{z}_{\mathbf{m}}$,
we perform early fusion by concatenating them along the channel dimension.
The fused latent representation is converted into transformer tokens through the patchification operator in Fig.~\ref{fig:overview},  illustrating:
\begin{align}
\mathbf{h}_t
=
\text{Patchify}\big([\mathbf{z}_t;\mathbf{z}_{\mathbf{m}}]\big),
\end{align}
where $[\cdot;\cdot]$ denotes channel concatenation and
$\mathbf{h}_t$ denotes the sequence of latent tokens processed by the DiT throughout the remainder of the paper.

Segmentation masks may correspond to different anatomical or pathological concepts.
To explicitly specify mask semantics, we introduce a compound text prompt noted $\mathbf{c}_\text{text}$.
A semantic pre-prompt describing the mask identity $\mathbf{c}_\mathbf{m}$ is concatenated before the optional radiology report $\mathbf{c}_r$.
When segmentation is absent, the pre-prompt is replaced with a sentence indicating an empty condition (\texttt{``no image segmentation is provided''}).
This design enables optional conditioning while preserving consistent text structure.

\vspace{3pt}
\vspace{3pt}
\noindent \textbf{Modified DiT Block.}
Our architecture builds upon the diffusion transformer (DiT)~\cite{Peebles2022DiT} but introduces several modifications to satisfy controllability and anatomical fidelity requirements. 
First, to incorporate token-level textual conditioning, we introduce a multi-head cross-attention (MHCA) module to DiT that attends image tokens to compound text tokens.
Adaptive layer normalization (AdaLN) is applied before and after MHCA to modulate features using timestep embeddings.
Second, to stabilize learning and improve anatomical fidelity, we introduce a depthwise separable convolution after each transformer block, enforcing local spatial consistency between neighboring tokens.
Third, to improve reasoning over long radiology reports and compound text, gated attention~\cite{qiu2025gated} is incorporated in every layer.
Ablation studies in Supp.~\ref{supp:gated} further prove the efficacy of each component in the aforementioned modifications.

Illustrated in Fig.~\ref{fig:overview}, let $\mathbf{h}^{j}$ denote tokens at layer $j$ and
$\mathbf{c}_{\text{text}}$ denote compound text tokens, our modified DiT block is summarized as:
\begin{equation}
\begin{aligned}
&\mathbf{Q}_h = \mathbf{h}^{j}\mathbf{W}_h^{Q},
\quad
\mathbf{K}_h = \mathbf{c}_{\text{text}}\mathbf{W}_h^{K},
\quad
\mathbf{V}_h = \mathbf{c}_{\text{text}}\mathbf{W}_h^{V},
\qquad
{\color{blue}\mathbf{G}_h = \mathbf{h}^{j}\mathbf{W}_h^{G}},\\[3pt]
&{\color{blue}\widetilde{\mathbf{A}}_h}
=
{\color{blue}\sigma(\mathbf{G}_h)}
\odot
\mathrm{Softmax}
\!\left(
\frac{\mathbf{Q}_h\mathbf{K}_h^\top}{\sqrt{d_h}}
\right)\mathbf{V}_h,\\[3pt]
&\mathbf{u}^j
=
{\color{blue}\mathrm{AdaLN}
\big(
\mathrm{MHCA}(\mathbf{h}^{j},\mathbf{c}_{\text{text}})
\big)}
+\mathbf{h}^{j},\\
&\mathbf{v}^j
=
\mathrm{AdaLN}
(
\mathrm{FFN}(\mathbf{u}^j)
)
+\mathbf{u}^j,\\
&\mathbf{h}^{j+1}
=
{\color{blue}\mathrm{Conv}}(\mathbf{v}^j),
\end{aligned}
\label{eq:model}
\end{equation}
where the $\text{Conv}(\cdot)$ in the last line involves a spatial and channel-wise convolution to enhance inter-patch consistency as shown in Fig.~\ref{fig:overview}  (ablation in Supp.~\ref{supp:gated}). 
For intra-patch learning, we adopt gated attention to multi-head self-attention (MHSA) block as well.
After all blocks are finished, We perform the unpatchify operation \textit{i.e.}, $\hat{\mathbf{z}}_0
=
\text{UnPatchify}\big(\mathbf{h}^J)$, on which we apply our loss function.

\subsection{Training and Inference Procedure}

We train the model using conditional rectified flow~\cite{liu2023flow, lipman2023flow} in the latent token space following the $x$-prediction formulation of~\cite{li2025back}.
Let $\mathbf{z}_0$ denote clean latent tokens obtained from the tokenizer and
$\boldsymbol{\epsilon}\sim\mathcal{N}(0,\mathbf{I})$ denote Gaussian noise.
A noisy sample is constructed using linear interpolation:
\begin{align}
\mathbf{z}_t
=
(1-t)\mathbf{z}_0 + t\boldsymbol{\epsilon},
\qquad t\in[0,1].
\end{align}

Our network directly predicts the clean latent tokens
$\mathbf{z}_\theta = f_\theta(\mathbf{z}_t,t,\mathbf{c})$, where $\mathbf{c}$ includes $\mathbf{c}_\mathbf{m}, (\mathbf{c}_r), \mathbf{z}_\mathbf{m}$.
Thus velocity prediction and conditional flow matching objective are computed as:
\begin{align}
v_\theta(\mathbf{z}_t,t,\mathbf{c})
:=
\frac{
\mathbf{z}_\theta - \mathbf{z}_t
}{
0-t
}; \quad
\mathcal{L}
=
\mathbb{E}_{t,\mathbf{z}_0,\boldsymbol{\epsilon}}
\Big[
\|
v_\theta(\mathbf{z}_t,t,\mathbf{c})
-
\frac{\mathbf{z}_0-\mathbf{z}_t}{0-t}
\|_2^2
\Big],
\label{eq:training}
\end{align}
which corresponds to $x$-prediction with $v$-loss as described in~\cite{li2025back}.
Directly predicting clean latent tokens encourages learning on the data manifold and improves anatomical fidelity.

\paragraph{Conditioning Randomization.}
During training, conditioning modalities are randomized within each batch to encourage robustness. Samples may include (i) radiology report only (i.e., $\mathbf{c}_\mathbf{m} = \emptyset$),
(ii) segmentation mask with semantic description only (i.e., $\mathbf{c}_r = \emptyset$),
(iii) both prompts, or (iv) neither prompt. This strategy enables the model to operate under arbitrary prompt availability during inference. 
We set segmentation tokens to zeros ($\mathbf{z}_{\mathbf{m}}=\mathbf{0}$) when segmentation masks are absent. 

\paragraph{Inference.}
Given the learned velocity field $v_\theta$, generation is performed by solving the ordinary differential equation (ODE):
\begin{align}
\frac{d\mathbf{z}_t}{dt}
=
v_\theta(\mathbf{z}_t,t,\mathbf{c}),
\end{align}
starting from $\mathbf{z}_1 \sim \mathcal{N}(0,\mathbf{I})$ and integrating to $t=0$.
We use a Dormand–Prince solver with a fixed 50 steps to solve the ODE.
After the last step of denoising, we pass the denoised $\mathbf{z}_0$ to our decoder to obtain the image, \textit{i.e.}, $\hat{\mathbf{x}} = \text{Dec}(\hat{\mathbf{z}}_0)$.
\section{Experiments}
We conduct  experiments to show \textbf{flexibility, controllability, scalability} and diversity of our model.
We compare our method with recent open-source SOTA methods with publicly available weights, including GenerateCT~\cite{generatect},     MedSyn~\cite{medsyn2024}, and Text2CT~\cite{molino2025texttoctgeneration3dlatent} (top candidate in VLM3D Challenge~\cite{Hamamci2024GeneralistFM}).
We compare results in text-only, mask-only, and joint-modality settings to show flexibility and controllability, different random seeds to show diversity, and two downstream tasks to show scalability.
Moreover, we provide radiologists' assessment to present close alignment of our results with real images. 
(Ablation study in Supp.~\ref{supp:radiology})

\subsection{Dataset}
We perform our training and validation on the CT-Rate dataset~\cite{Hamamci2024GeneralistFM}, a large public chest CT dataset consisting of 25,692 pairs of CT images and radiology reports, with the corresponding 18 abnormality labels. 
We follow their official train-validation setting.
We follow MedSyn~\cite{medsyn2024} and affine-register all CT images, with a fixed pixel spacing $0.7\times0.7\times0.7$ followed by padding or cropping to a fixed size $448\times448\times448$. 
Hounsfield Unit (HU) intensities are clipped to $[-1024, 1024]$ and normalized to $[-1, 1]$ for stabilized training~\cite{molino2025texttoctgeneration3dlatent}. 
CT-Rate does not contain corresponding segmentation masks, thus we leverage multiple state-of-the-art models for anatomical and pathological masks. 
We include four anatomy and five abnormality masks: lung lobes (lungmask~\cite{lungmask}), airways (TotalSegmentator~\cite{TotalSegmentator}), vessels (\cite{POLETTIVessel}), heart (TotalSegmentator~\cite{TotalSegmentator}); consolidation (MedPSeg~\cite{carmo2024medpseg}), ground glass opacity (MedPSeg~\cite{carmo2024medpseg}), pericardial effusion (~\cite{pericardial}), pleural effusion (~\cite{pleuralEffu}) and lung nodules (TotalSegmentator~\cite{TotalSegmentator}).

\subsection{Experimental details}
\textbf{Textual Conditions.} We encode the compound pre-prompt $\mathbf{c}_\mathbf{m}$ and the radiology reports $\mathbf{c}_r$ separately to reduce storage space.
Both the radiology reports (including both Findings and Impressions) and the compound pre-prompt are processed with a pretrained CXR-BERT~\cite{cxrbert} model from MedSyn~\cite{medsyn2024}.
For training data variance, we augment $\mathbf{c}_\mathbf{m}$ using different formats shown in Supp.~\ref{supp:paraphrase}.
We set the $\mathbf{c}_\mathbf{m}$ max token length as $32$, describing identity of abnormalities.
For the optional reports, we concatenate the findings and impressions with a \texttt{<SEP>} token in between, and set token length $512$. 
Thus, we acquire textual condition shaped $(544, 768)$ concatenating them. (More details in Supp.~\ref{supp:imple})

\noindent \textbf{Mask Conditions and Tokens.} 
We normalize all masks to $[-0.5, 0.5]$ for matching our image intensity and avoiding outliers.
For all images and segmentation masks shaped $448\times448\times448$, our encoder processes them into latents $\mathbf{z}$ of a unified shape $(n_d \times n_h \times n_w) \times n_c = (112, 56, 56)\times4$. 
We z-norm all latents with their individual channels for training with the loss fucntion in Eq.~\ref{eq:training}, and reverse the z-norm when applying our decoder.
(More details in Supp.~\ref{supp:imple})

\subsection{Quantitative Results}

\begin{table*}[t!]
\caption{\textbf{FID$\downarrow$ on different CT windows with reports only.}
We use different intensity windows used by radiologists to increase contrast for different organs: ([Level, Width]= Lung: $[-600, 1500]$, Vessel: $[100, 700]$, Soft Tissue: $[50, 350]$, Bone: $[300, 2000]$) when comparing FID scores. 
(best \textbf{bold}, second  \underline{underlined})
Our method performs best over all windows on almost all axes. 
(XY: Axial, YZ: Sagittal, ZX: Coronal)
}
\begin{adjustbox}{max width=\textwidth, center}
\begin{tabular}{lcccccccccccccccc}
\toprule
 & \multicolumn{4}{c}{\textbf{Lung}} 
 & \multicolumn{4}{c}{\textbf{Vessel}}
 & \multicolumn{4}{c}{\textbf{Soft Tissue}}
 & \multicolumn{4}{c}{\textbf{Bone}} \\
\cmidrule(lr){2-5}\cmidrule(lr){6-9}
\cmidrule(lr){10-13}\cmidrule(lr){14-17}

\textbf{Model} &
XY$\downarrow$ & YZ$\downarrow$ & ZX$\downarrow$ & Mean$\downarrow$ &
XY$\downarrow$ & YZ$\downarrow$ & ZX$\downarrow$ & Mean$\downarrow$ &
XY$\downarrow$ & YZ$\downarrow$ & ZX$\downarrow$ & Mean$\downarrow$ &
XY$\downarrow$ & YZ$\downarrow$ & ZX$\downarrow$ & Mean$\downarrow$ \\
\midrule
GenerateCT~\cite{generatect} &
$\mathbf{7.03}$ & $\underline{14.07}$ & $21.71$ & $14.27$ &
$\mathbf{4.78}$ & $11.77$ & $14.57$ & $10.37$ &
$\mathbf{6.49}$ & $13.89$ & $17.25$ & $12.54$ &
$\mathbf{4.57}$ & $9.20$ & $12.35$ & $8.71$ \\

MedSyn~\cite{medsyn2024} &
$\underline{9.37}$ & $14.27$ & $\underline{14.84}$ & $12.83$ &
$13.59$ & $16.33$ & $17.24$ & $15.72$ &
$13.96$ & $18.59$ & $20.00$ & $17.52$ &
$\underline{5.86}$ & $8.23$ & $8.22$ & $7.44$ \\

Text2CT~\cite{molino2025texttoctgeneration3dlatent} &
$11.51$ & $\underline{8.55}$ & $17.69$ & $\underline{12.58}$ &
$8.91$ & $\underline{9.76}$ & $\underline{9.37}$ & $\underline{9.35}$ &
$9.12$ & $\underline{13.18}$ & $\underline{13.58}$ & $\underline{11.96}$ &
$8.67$ & $\underline{7.04}$ & $\underline{6.37}$ & $\underline{7.36}$ \\

\ours (Ours) &
$10.96$ & $\mathbf{6.40}$ & $\mathbf{11.40}$ & $\mathbf{9.59}$ &
$\underline{7.97}$ & $\mathbf{5.17}$ & $\mathbf{6.24}$ & $\mathbf{6.46}$ &
$\underline{8.97}$ & $\mathbf{7.45}$ & $\mathbf{8.99}$ & $\mathbf{8.47}$ &
$7.13$ & $\mathbf{3.43}$ & $\mathbf{3.19}$ & $\mathbf{4.58}$ \\

\bottomrule
\end{tabular}
\end{adjustbox}
\label{tab:fid_all}
\end{table*}

\noindent \textbf{Modified DiT.} 
We choose patch size $(2\times 2\times 2)$ to balance between memory issues and image quality, yielding $43,904$ patches (latent tokens). 
We use $16$ layers and feature dimension $1152$~\cite{Peebles2022DiT}. 
All the experiments are run on a single A100 GPU (PyTorch 2.6.0+cu126) with AdamW optimizer, learning rate $10^{-4}$ and batch size $1$ with $4$ accumulated forward steps.
(More details in Supp.~\ref{supp:imple})


\noindent \textbf{Baselines and Metrics.} 
We compare on various metrics against current state-of-the-art text-to-image generative models, namely GenerateCT~\cite{generatect}, MedSyn~\cite{medsyn2024}, Text2CT~\cite{molino2025texttoctgeneration3dlatent}. 
For perceptual evaluation, we follow radiology practice and define contrast intensity windows for four tissue types: lung, vessel, soft tissue, and bone (See Supp.~\ref{supp:imple}). 
We compute FID for each window across the three orthogonal planes.
For semantic perspectives, we leverage two frequently used CLIP models: CT-CLIP~\cite{Hamamci2024GeneralistFM} and Pillar-0~\cite{pillar0} to compute CLIP-Score on both text-to-image (T2I) and image-to-image (I2I) scopes.
Note Pillar-0 applies eleven windows in their image encoder.
For factual correctness, we leverage the zero-shot CT-CLIP~\cite{Hamamci2024GeneralistFM} to predict the $18$ pathological labels based on the generated images. 
We report accuracy, precision, AUROC and F1 score based on the ground truth.

\begin{table*}[t!]
\centering

\begin{minipage}[t]{0.48\linewidth}
\centering
\caption{\textbf{CLIP-Score (\%) using reports only} with CT-CLIP ($1$ window) and Pillar-0 ($11$ windows). 
Our result achieves good results on these four metrics with both CLIP models. 
}
\begin{adjustbox}{max width=\linewidth, center}
\begin{tabular}{lcccc}
\toprule
\multirow{2}{*}{\textbf{Source}} 
& \multicolumn{2}{c}{\textbf{CT-CLIP} ($1$ window)} 
& \multicolumn{2}{c}{\textbf{Pillar-0} ($11$ windows)} \\
\cmidrule(lr){2-3} \cmidrule(lr){4-5}
& \textbf{T2I} $\uparrow$ & \textbf{I2I} $\uparrow$
& \textbf{T2I} $\uparrow$ & \textbf{I2I} $\uparrow$ \\
\midrule
Real 
& 16.46 & -- 
& 26.36 & -- \\
\midrule

GenerateCT~\cite{generatect}
& 14.70 & 59.03
& 22.10 & 32.11 \\

MedSyn~\cite{medsyn2024}
& 12.70 & \underline{81.61}
& 22.75 & 44.59 \\

Text2CT~\cite{molino2025texttoctgeneration3dlatent}
& $\mathbf{20.17}$ & 72.37
& $\underline{25.80}$ & $\mathbf{49.38}$ \\

\ours (Ours)
& $\underline{14.74}$ & $\mathbf{83.00}$
& $\mathbf{27.61}$ & $\underline{46.79}$ \\
\bottomrule
\end{tabular}
\end{adjustbox}
\label{tab:clip_comparison}
\end{minipage}
\hfill
\begin{minipage}[t]{0.48\linewidth}
\centering
\caption{\textbf{Classification results (\%) using reports only}. We use zeroshot CT-CLIP as classifier, showing our method performs on par with existing method. 
}
\begin{adjustbox}{max width=\linewidth, center}
\begin{tabular}{lcccc}
\toprule
& \textbf{Accuracy} $\uparrow$ & \textbf{Precision} $\uparrow$
& \textbf{AUROC} $\uparrow$ & \textbf{F1} $\uparrow$ \\
\midrule
Real 
 & $67.81$ & $32.00$ & ${67.34}$ & ${58.95}$\\

\midrule
GenerateCT~\cite{generatect}
& $28.24$ & $19.77$
&  $50.57$ & $26.33$\\

MedSyn~\cite{medsyn2024}
& $\mathbf{79.65}$ & $\mathbf{35.65}$
& $\underline{51.32}$ & $\underline{40.12}$ \\

Text2CT~\cite{molino2025texttoctgeneration3dlatent}
& $45.06$ & $23.76$
& $50.96$ & $37.43$ \\

\ours (Ours)
& $\underline{63.55}$ & $\underline{24.90}$
& $\mathbf{52.24}$ & $\mathbf{48.40}$ \\
\bottomrule
\end{tabular}
\end{adjustbox}
\label{tab:classifications}
\end{minipage}
\end{table*}

\begin{figure}
    \centering
    \includegraphics[width=1.0\linewidth]{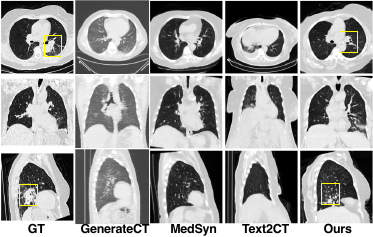}
    \caption{
    \textbf{Comparisons using text guidance only.} 
    All images are shown in lung contrast window with pixel spacing $(0.7\times 0.7 \times0.7)$.
    Corresponding report mentions \underline{`volume loss in the left lung and widespread atelectatic changes'}. 
    Our method correctly synthesizes this in bounding box, showing \textbf{controllability}.
    }
    \label{fig:res_text}
\end{figure}


\begin{table}[t!]
\centering

\begin{minipage}{0.34\linewidth}
\centering
\small
\captionof{table}{\textbf{Expert evaluation} of images matching report and mask.}
\label{tab:human_eval}
\begin{adjustbox}{max width=\linewidth}
\begin{tabular}{lccc}
\toprule
 & Match & \makecell{Partial\\Match} & \makecell{Not\\Match} \\

\midrule
\multicolumn{4}{l}{Radiologist 1 (4 years experience)} \\
Real & 49 & 10 & 16  \\
Synthetic & 40 & 18 & 17  \\
\midrule
\multicolumn{4}{l}{Radiologist 2 (17 years experience)} \\
Real & 45 & 25 & 5  \\
Synthetic & 36 & 30 & 9  \\
\bottomrule
\end{tabular}
\end{adjustbox}
\end{minipage}
\hfill
\begin{minipage}{0.63\linewidth}
\centering
\small
\captionof{table}{
\textbf{Results of incorporating synthesized data for Opacity segmentation on SemiSeg test set.}
We report results with the real images, our synthesized images, and their combination. (\textbf{best}, \underline{second best}) }
\label{tab:covid_results}
\begin{adjustbox}{max width=\linewidth}
\begin{tabular}{lcccc}
\toprule
\textbf{Training Data} & \textbf{Dice$(\%)\uparrow$} & \textbf{Jaccard$(\%)\uparrow$} & \textbf{ASD (pixel)$\downarrow$} & \textbf{HD95 (pixel)$\downarrow$} \\
\midrule
Real (50 images) & $\underline{58.78}$ & $\underline{43.49}$ & $\mathbf{21.65}$ & $65.33$ \\
Synthesized (416 images) & $48.82$ & $32.30$ & $56.32$ & $\underline{10.61}$ \\
Combined (466 images) & $\mathbf{68.13}$ & $\mathbf{54.27}$ & $\underline{43.32}$ & $\mathbf{4.69}$ \\
\midrule
Inf-Net~\cite{fan2020infnet} & $62.41$ & $45.35$ & $32.67$ & $20.63$ \\
MedPSeg~\cite{carmo2024medpseg} & $65.20$ & $48.13$ & $28.45$ & $5.51$ \\
\bottomrule
\end{tabular}
\end{adjustbox}
\end{minipage}

\end{table}


\section{Results}

\subsection{Quantitative Results}

\textbf{Comparing Baselines.} Table~\ref{tab:fid_all} demonstrates that our method outperforms all baselines in the mean FID scores among all contrast windows.
We achieve $\mathbf{24\%}$ improvement compared to TextCT on the lung window (ours $\mathbf{9.59}$ \textit{vs.} TextCT $\mathbf{12.58}$).
The good performances result from our gated attention, which captures most details in long CT reports.
GenerateCT performs good on axial (XY) view, due to its design of slice-wise super resolution on axial views.
It however loses continuity and suffer on the other two axes.
MedSyn suffers from soft tissue generation due to its training on another dataset, while Text2CT uses unregistered images for training, yielding to inferior results.
Likewise, tables~\ref{tab:clip_comparison}~\ref{tab:classifications} also substantiate that our method performs well on CLIP T2I, I2I, factual correctness AUROC and F1 scores (table~\ref{tab:clip_comparison} Pillar-0 T2I ours $\mathbf{27.6\%}$ \textit{vs.} Text2CT $\mathbf{25.8\%}$).
This results from our modified DiT-based method, as it enhances report understanding by token level cross attention, with inter-patch consistency.

\noindent \textbf{Radiologist Evaluation.}
We randomly sample $15$ pairs each abnormality, and synthesize another $75$ cases with reports and masks, forming a dataset totaling $150$ cases. 
A double-blind test with two radiologists to evaluate whether the images match the report and mask is conduct. 
Table~\ref{tab:human_eval} shows the assessment of synthetic ones and real ones. 
Our synthetic images show majority in the `Match' and `Partial Match' ones, with a distribution similar to the real ones.
Moreover, both radiologists show agreement on low `Not Match' counts, further implying the same pattern.
The radiologist evaluation shows an overall high fidelity result of our image towards the input condition.
(More details in Supp.~\ref{supp:radiology})

\subsection{Qualitative Results}

\begin{figure}
    \centering
    \includegraphics[width=1.0\linewidth]{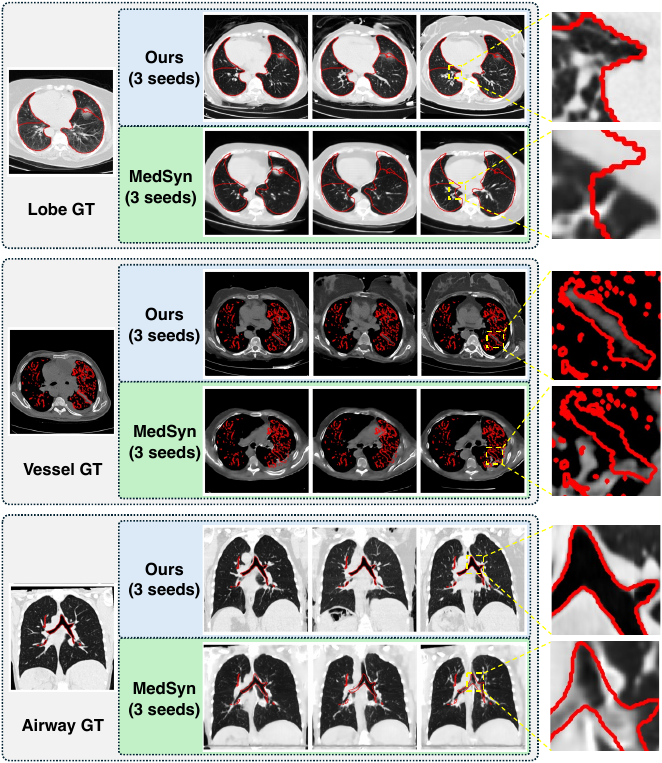}
    \caption{
    \textbf{Comparison of anatomically conditional generation using lobe, airway and vessel masks (without report).} 
    The second block is shown in vessel contrast window, while the other two in lung contrast window.
    We show ground truth images, ours and MedSyn's with 3 different random seeds. 
    The zoomed-in views on the right show MedSyn fails to follow anatomical masks.
    In contrast, our results closely respect the conditional masks, exhibiting both \textbf{controllability} and \textbf{diversity}.
    }
    \label{fig:res_seg_medsyn}
\end{figure}

\begin{figure}[t!]
    \centering
    \includegraphics[width=1.0\linewidth]{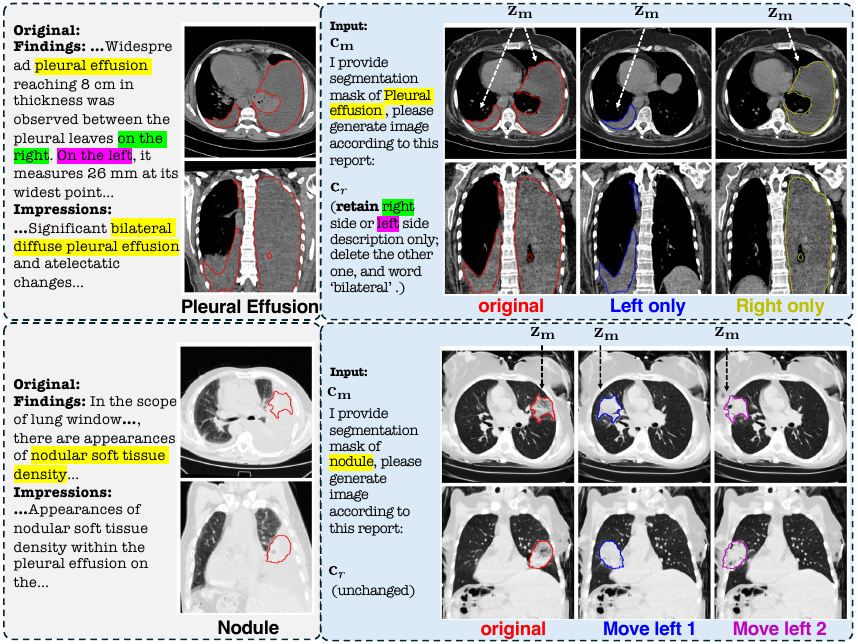}
    \caption{
    \textbf{Results of modifying pathology masks and reports.} 
    For pleural effusion, we \textbf{retain} either \textcolor{blue}{left} or \textcolor{green}{right} mask, and modify the corresponding report by deleting the abnormality description on the other side.
    For nodule, since no location was described in report, we spatially \textbf{move} the mask to left side of body by either \textcolor{blue}{flipping} or \textcolor[HTML]{BF01C0}{translating} without changing the report. 
    Our results closely follow the conditional masks, showing both \textbf{controllability} and \textbf{flexibility}.
    }
    \label{fig:res_modify_seg}
\end{figure}


\textbf{Comparing Baselines.} In Fig.~\ref{fig:res_text}, although GenerateCT shows good on axial view (first row), it suffers on the other two, as it performs super resolution on axial slices. 
MedSyn generates a rather healthy image, lacking the focus on the abnormality region.
Text2CT shows abdomen which leads to good soft tissues yet it produces blurry image compared to ours, since it adopts un-registered image and therefore loses detail. 
In contrast, our method generates high quality images, while it clearly synthesizes the volume loss and atelectasis (yellow boxes) noted in the report, attributing to our encoding of word level tokens and gated attention module, allowing for more details preserved.
In Fig.~\ref{fig:res_seg_medsyn}, our method shows closer alignment towards the segmentation mask. 
As noted by the zoomed-in views (right most), MedSyn generates organs falling out of the given mask, which results from the loose incorporation of segmentation masks.
Ours on the contrary, directly incorporates the noisy latent and mask together, enforces better inter-patch consistency with our convolution layer, therefore yielding better alignment.
Results in Fig.~\ref{fig:res_modify_seg} show great controllability over masks of different appearance.
Whether we retain one side of the abnormality, or completely flip/translate the mask, the generated results always show strong correspondence with the condition.
All these results indicate our controllability and flexibility.
(More results in Supp.~\ref{supp:mask})

\subsection{Scalability}
We investigate the scalability with two downstream tasks: 
1) augmenting datasets assisting training segmentation model.
2) Multiple conditional synthesis simulating gradually growing abnormalities.
(More results in Supp.~\ref{supp:augment})

\noindent \textbf{Augmenting dataset.}
We augment the SemiSeg dataset~\cite{fan2020infnet} segmenting Ground Glass Opacity (GGO), whose original training set contains only $50$ 2D images with segmentation labels. 
We sample $416$ 2D masks from 3D ones (details in Supp.~\ref{supp:imple})
and compare three settings: 1) real images only (50 images), 2) synthesized only (416 images), and 3) combination of both (466 images).
We train vanilla UNet models and show quantitative results in table~\ref{tab:covid_results}.
Training with the combined dataset achieves the highest Dice score ($\mathbf{68.13\%}$), Jaccard ($\mathbf{54.27\%}$), and the lowest HD95 ($\mathbf{4.69}$), even surpassing the SOTA models (MedPSeg Dice $\mathbf{65.2\%}$), showing great efficacy with our augmented images.
In addition, training solely on synthetic images achieves a fair Dice score (${48.82\%}$) and HD95 (${10.61}$).
In conclusion, 1) our method effectively assists training segmentation models by augmenting training data. 
2) our model receives only segmentation masks and preserves data and patient privacy.  
(More results in Supp.~\ref{supp:augment})

\begin{figure}[t!]
    \centering
    \includegraphics[width=0.8\linewidth]{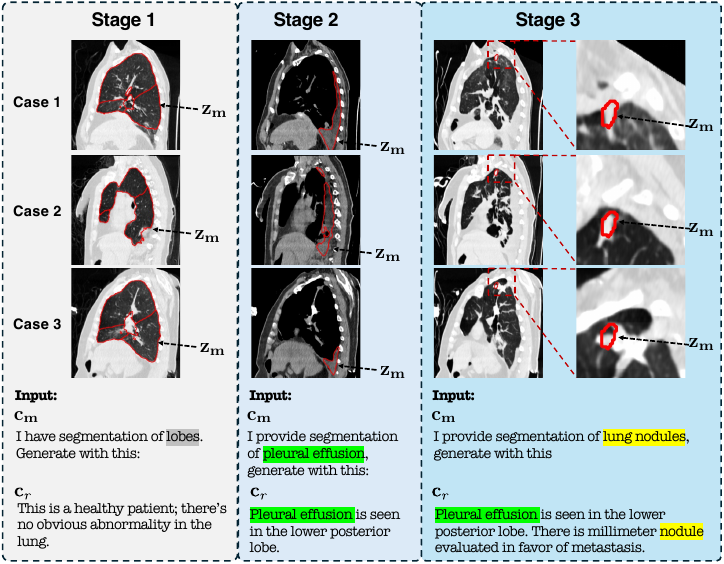}
    \caption{
    \textbf{Results of progressively adding conditions.} 
    We start by generating healthy images with lobes only. 
    Progressively, we add a pleural effusion mask and a nodule mask, describing them in the radiology report.
    In each row, following stages keep the characteristic from previous stages, while exhibiting new abnormalities.
    }
    \label{fig:res_scalability}
    \vspace{-0.5em}
\end{figure}

\noindent \textbf{Multiple Conditional Synthesis.}
Fig.~\ref{fig:res_scalability} shows the results where we progressively add more details to a case, synthesizing growing multiple diseases. 
We start with lobe mask and healthy report (pair one) to generate stage one images, then fix the same random seed and add noise again to apply additional pairs of conditions following SDEdit~\cite{meng2022sdedit}.
Images in column two all preserve the original layout of the first column, with only pleural effusions added and marked in \textcolor{red}{boundaries}. 
Similarly in column three, while lobe positions and pleural effusions are preserved, the zoomed-in views show delicate nodules, indicating our scalability of multiple conditions.
In conclusion,
1) scalability of our model indicates further details can be added given single or more segmentation masks with our low-cost controlling ,
2) the results form procedures with arbitrary abnormalities and simple reports, educating residents to various combinations of diseases on the same CT.

\section{Conclusion}
This paper proposes the first joint-modality to lung CT generation model enabling single object of anatomy or pathology as condition, enforcing scalability and flexibility via compound prompts (pre-prompt and radiology reports). 

%
%

\clearpage 


\begin{thebibliography}{10}
\providecommand{\url}[1]{\texttt{#1}}
\providecommand{\urlprefix}{URL }
\providecommand{\doi}[1]{https://doi.org/#1}

\bibitem{amirrajab2025radiologyreportconditional3d}
Amirrajab, S., Salahuddin, Z., Kuang, S., Woodruff, H.C., Lambin, P.: Radiology report conditional 3d ct generation with multi encoder latent diffusion model (2025), \url{https://arxiv.org/abs/2509.14780}

\bibitem{cxrbert}
Boecking, B., Usuyama, N., Bannur, S., Castro, D.C., Schwaighofer, A., Hyland, S., Wetscherek, M., Naumann, T., Nori, A., Alvarez-Valle, J., Poon, H., Oktay, O.: Making the most of text semantics to improve biomedical vision-language processing (2022). \doi{10.48550/ARXIV.2204.09817}, \url{https://arxiv.org/abs/2204.09817}

\bibitem{carmo2024medpseg}
Carmo, D.S., Ribeiro, J.A., Comellas, A.P., Reinhardt, J.M., Gerard, S.E., Rittner, L., Lotufo, R.A.: Medpseg: Hierarchical polymorphic multitask learning for the segmentation of ground-glass opacities, consolidation, and pulmonary structures on computed tomography (2024)

\bibitem{chung2023solving}
Chung, H., Ryu, D., McCann, M.T., Klasky, M.L., Ye, J.C.: Solving 3d inverse problems using pre-trained 2d diffusion models (2023)

\bibitem{medddpm}
Dorjsembe, Z., Pao, H.K., Odonchimed, S., Xiao, F.: Conditional diffusion models for semantic 3d brain mri synthesis. IEEE Journal of Biomedical and Health Informatics  \textbf{28}(7),  4084--4093 (2024). \doi{10.1109/JBHI.2024.3385504}

\bibitem{fan2020infnet}
Fan, D.P., Zhou, T., Ji, G.P., Zhou, Y., Chen, G., Fu, H., Shen, J., Shao, L.: Inf-net: Automatic covid-19 lung infection segmentation from ct images. IEEE Transactions on Medical Imaging  \textbf{39}(8),  2626--2637 (2020). \doi{10.1109/TMI.2020.2996645}

\bibitem{text2CTGuo}
Guo, P., Zhao, C., Yang, D., He, Y., Nath, V., Xu, Z., Bassi, P., Zhou, Z., Simon, B., Harmon, S., Turkbey, B., Xu, D.: Text2ct: Towards 3d ct volume generation from free-text descriptions using diffusion model (05 2025). \doi{10.48550/arXiv.2505.04522}

\bibitem{maisiv1}
Guo, P., Zhao, C., Yang, D., Xu, Z., Nath, V., Tang, Y., Simon, B., Belue, M., Harmon, S., Turkbey, B., Xu, D.: Maisi: Medical ai for synthetic imaging (09 2024). \doi{10.48550/arXiv.2409.11169}

\bibitem{Hamamci2024GeneralistFM}
Hamamci, I.E., Er, S., Almas, F., Simsek, A.G., Esirgun, S.N., İrem Hatice~Doğan, Dasdelen, M.F., Wittmann, B., Simsar, E., Simsar, M., Erdemir, E.B., Alanbay, A., Sekuboyina, A.K., Lafci, B., Ozdemir, M.K., Menze, B.H.: Generalist foundation models from a multimodal dataset for 3d computed tomography. Nature biomedical engineering  (2024), \url{https://api.semanticscholar.org/CorpusID:268691507}

\bibitem{generatect}
Hamamci, I.E., Er, S., Sekuboyina, A., Simsar, E., Tezcan, A., Simsek, A.G., Esirgun, S.N., Almas, F., Do\u{g}an, I., Dasdelen, M.F., Prabhakar, C., Reynaud, H., Pati, S., Bluethgen, C., Ozdemir, M.K., Menze, B.: Generatect: Text-conditional generation of 3d chest ct volumes. In: Computer Vision – ECCV 2024: 18th European Conference, Milan, Italy, September 29–October 4, 2024, Proceedings, Part LXXIX. p. 126–143. Springer-Verlag, Berlin, Heidelberg (2024). \doi{10.1007/978-3-031-72986-7_8}, \url{https://doi.org/10.1007/978-3-031-72986-7_8}

\bibitem{HanMedGen3D}
Han, K., Xiong, Y., You, C., Khosravi, P., Sun, S., Yan, X., Duncan, J.S., Xie, X.: Medgen3d: A deep generative framework for paired 3d image and mask generation. In: Greenspan, H., Madabhushi, A., Mousavi, P., Salcudean, S., Duncan, J., Syeda-Mahmood, T., Taylor, R. (eds.) Medical Image Computing and Computer Assisted Intervention -- MICCAI 2023. pp. 759--769. Springer Nature Switzerland, Cham (2023)

\bibitem{lungmask}
Hofmanninger, J., Prayer, F., Pan, J., Röhrich, S., Prosch, H., Langs, G.: Automatic lung segmentation in routine imaging is primarily a data diversity problem, not a methodology problem. European Radiology Experimental  \textbf{4}, ~50 (08 2020). \doi{10.1186/s41747-020-00173-2}

\bibitem{lungddpm}
Jiang, Y., Lemaréchal, Y., Plante, S., Bafaro, J., Abi-Rjeile, J., Joubert, P., Després, P., Manem, V.: Lung-ddpm: Semantic layout-guided diffusion models for thoracic ct image synthesis. IEEE Transactions on Biomedical Engineering  \textbf{73}(3),  1134--1145 (2026). \doi{10.1109/TBME.2025.3599011}

\bibitem{Krishna_2025}
Krishna, A., Wang, G., Mueller, K.: Guided synthesis of annotated lung ct images with pathologies using a multi-conditioned denoising diffusion probabilistic model (mddpm). Physics in Medicine \& Biology  \textbf{70}(6),  065007 (mar 2025). \doi{10.1088/1361-6560/adb9b3}, \url{https://doi.org/10.1088/1361-6560/adb9b3}

\bibitem{kumar2025prism}
Kumar, A., Kriz, A., Havaei, M., Arbel, T.: {PRISM}: High-resolution \& precise counterfactual medical image generation using language-guided stable diffusion. In: Medical Imaging with Deep Learning (2025), \url{https://openreview.net/forum?id=UpJMAlZNuo}

\bibitem{li2025back}
Li, T., He, K.: Back to basics: Let denoising generative models denoise. arXiv preprint arXiv:2511.13720  (2025)

\bibitem{lin2024opensoraplanopensourcelarge}
Lin, B., Ge, Y., Cheng, X., Li, Z., Zhu, B., Wang, S., He, X., Ye, Y., Yuan, S., Chen, L., Jia, T., Zhang, J., Tang, Z., Pang, Y., She, B., Yan, C., Hu, Z., Dong, X., Chen, L., Pan, Z., Zhou, X., Dong, S., Tian, Y., Yuan, L.: Open-sora plan: Open-source large video generation model (2024), \url{https://arxiv.org/abs/2412.00131}

\bibitem{lipman2023flow}
Lipman, Y., Chen, R.T.Q., Ben-Hamu, H., Nickel, M., Le, M.: Flow matching for generative modeling. In: The Eleventh International Conference on Learning Representations (2023), \url{https://openreview.net/forum?id=PqvMRDCJT9t}

\bibitem{Liu_2024_ACCV}
Liu, C., Yuan, X., Yu, Z., Wang, Y.: Texdc: Text-driven disease-aware 4d cardiac cine mri images generation. In: Proceedings of the Asian Conference on Computer Vision (ACCV). pp. 3005--3021 (December 2024)

\bibitem{liu2023flow}
Liu, X., Gong, C., qiang liu: Flow straight and fast: Learning to generate and transfer data with rectified flow. In: The Eleventh International Conference on Learning Representations (2023), \url{https://openreview.net/forum?id=XVjTT1nw5z}

\bibitem{radimagenet}
Mei, X., Liu, Z., Robson, P.M., Marinelli, B., Huang, M., Doshi, A., Jacobi, A., Cao, C., Link, K.E., Yang, T., Wang, Y., Greenspan, H., Deyer, T., Fayad, Z.A., Yang, Y.: Radimagenet: An open radiologic deep learning research dataset for effective transfer learning. Radiology: Artificial Intelligence  \textbf{0}(ja),  e210315 (0). \doi{10.1148/ryai.210315}, \url{https://doi.org/10.1148/ryai.210315}

\bibitem{meng2022sdedit}
Meng, C., He, Y., Song, Y., Song, J., Wu, J., Zhu, J.Y., Ermon, S.: {SDE}dit: Guided image synthesis and editing with stochastic differential equations. In: International Conference on Learning Representations (2022), \url{https://openreview.net/forum?id=aBsCjcPu_tE}

\bibitem{molino2025texttoctgeneration3dlatent}
Molino, D., Caruso, C.M., Ruffini, F., Soda, P., Guarrasi, V.: Text-to-ct generation via 3d latent diffusion model with contrastive vision-language pretraining (2025), \url{https://arxiv.org/abs/2506.00633}

\bibitem{katelyn}
Morrison, K., Mathur, A., Bradshaw, A., Wartmann, T., Lundi, S., Zandifar, A., Dai, W., Batmanghelich, K., Eslami, M., Perer, A.: A human-centered approach to identifying promises, risks, \& challenges of text-to-image generative ai in radiology (07 2025). \doi{10.48550/arXiv.2507.16207}

\bibitem{pillar0}
Nercessian, M., Agrawal, K., Liu, L., Lian, L., Harguindeguy, N., Wu, Y., Mikhael, P., Lin, G., Sequist, L., Fintelmann, F., Darrell, T., Bai, Y., Chung, M., Yala, A.: Pillar-0: A new frontier for radiology foundation models (11 2025). \doi{10.21203/rs.3.rs-8196619/v1}

\bibitem{oliveras2025land}
Oliveras, A., Mar{\'\i}, R., Redondo, R., Guardi{\`a}-Olivella, O., Tost, A., Nagarajan, B., Migliorelli, C., Ribas, V., Radeva, P.: {LAND}: Lung and nodule diffusion for 3d chest {CT} synthesis with anatomical guidance. In: Submitted to Medical Imaging meets EurIPS: MedEurIPS 2025 (2025), \url{https://openreview.net/forum?id=VTQwlZLq0a}, under review

\bibitem{Peebles2022DiT}
Peebles, W., Xie, S.: Scalable diffusion models with transformers. arXiv preprint arXiv:2212.09748  (2022)

\bibitem{opensora2}
Peng, X., Zheng, Z., Shen, C., Young, T., Guo, X., Wang, B., Xu, H., Liu, H., Jiang, M., Li, W., Wang, Y., Ye, A., Ren, G., Ma, Q., Liang, W., Lian, X., Wu, X., Zhong, Y., Li, Z., Gong, C., Lei, G., Cheng, L., Zhang, L., Li, M., Zhang, R., Hu, S., Huang, S., Wang, X., Zhao, Y., Wang, Y., Wei, Z., You, Y.: Open-sora 2.0: Training a commercial-level video generation model in $200k$. arXiv preprint arXiv:2503.09642  (2025)

\bibitem{POLETTIVessel}
Poletti, J., Bach, M., Yang, S., Sexauer, R., Stieltjes, B., Rotzinger, D.C., Bremerich, J., {Walter Sauter}, A., Weikert, T.: Automated lung vessel segmentation reveals blood vessel volume redistribution in viral pneumonia. European Journal of Radiology  \textbf{150},  110259 (2022). \doi{10.1016/j.ejrad.2022.110259}

\bibitem{qiu2025gated}
Qiu, Z., Wang, Z., Zheng, B., Huang, Z., Wen, K., Yang, S., Men, R., Yu, L., Huang, F., Huang, S., Liu, D., Zhou, J., Lin, J.: Gated attention for large language models: Non-linearity, sparsity, and attention-sink-free. In: The Thirty-ninth Annual Conference on Neural Information Processing Systems (2025), \url{https://openreview.net/forum?id=1b7whO4SfY}

\bibitem{rombach2021highresolution}
Rombach, R., Blattmann, A., Lorenz, D., Esser, P., Ommer, B.: High-resolution image synthesis with latent diffusion models (2021)

\bibitem{pleuralEffu}
Sexauer, R., Yang, S., Weikert, T., Poletti, J., Bremerich, J., Roth, J., Sauter, A., Anastasopoulos, C.: Automated detection, segmentation, and classification of pleural effusion from computed tomography scans using machine learning. Investigative Radiology  \textbf{Publish Ahead of Print} (03 2022). \doi{10.1097/RLI.0000000000000869}

\bibitem{TRACE}
Shao, M., Miao, X., Duan, H., Wang, Z., Chen, J., Huang, Y., Wu, X., Deng, J., Long, Y., Zheng, Y.: Trace: Temporally reliable anatomically-conditioned 3d ct generation with enhanced efficiency. In: Gee, J.C., Alexander, D.C., Hong, J., Iglesias, J.E., Sudre, C.H., Venkataraman, A., Golland, P., Kim, J.H., Park, J. (eds.) Medical Image Computing and Computer Assisted Intervention -- MICCAI 2025. pp. 627--637. Springer Nature Switzerland, Cham (2026)

\bibitem{Wang20243DMA}
Wang, H., Liu, Z., Sun, K., Wang, X., Shen, D., Cui, Z.: 3d meddiffusion: A 3d medical latent diffusion model for controllable and high-quality medical image generation. IEEE Transactions on Medical Imaging  \textbf{44},  4960--4972 (2024), \url{https://api.semanticscholar.org/CorpusID:274789446}

\bibitem{wang2025ctflowvideoinspiredlatentflow}
Wang, J., Reynaud, H., Erick, F.X., Kainz, B.: Ctflow: Video-inspired latent flow matching for 3d ct synthesis (2025), \url{https://arxiv.org/abs/2508.12900}

\bibitem{wang2025lavin}
Wang, Z., Xia, X., Chen, R., Yu, D., Wang, C., Gong, M., Liu, T.: Lavin-dit: Large vision diffusion transformer. In: Proceedings of the Computer Vision and Pattern Recognition Conference. pp. 20060--20070 (2025)

\bibitem{TotalSegmentator}
Wasserthal, J., Breit, H.C., Meyer, M.T., Pradella, M., Hinck, D., Sauter, A.W., Heye, T., Boll, D.T., Cyriac, J., Yang, S., Bach, M., Segeroth, M.: Totalsegmentator: Robust segmentation of 104 anatomic structures in ct images. Radiology: Artificial Intelligence  \textbf{5}(5),  e230024 (2023). \doi{10.1148/ryai.230024}, \url{https://doi.org/10.1148/ryai.230024}

\bibitem{pericardial}
Wilder-Smith, A.J., Yang, S., Weikert, T., Bremerich, J., Haaf, P., Segeroth, M., Ebert, L.C., Sauter, A., Sexauer, R.: Automated detection, segmentation, and classification of pericardial effusions on chest ct using a deep convolutional neural network. Diagnostics  \textbf{12}(5) (2022). \doi{10.3390/diagnostics12051045}, \url{https://www.mdpi.com/2075-4418/12/5/1045}

\bibitem{xing2024deep}
Xing, X., Ning, J., Nan, Y., Yang, G.: Deep generative models unveil patterns in medical images through vision-language conditioning. arXiv preprint arXiv:2410.13823  (2024)

\bibitem{medsyn2024}
Xu, Y., Sun, L., Peng, W., Jia, S., Morrison, K., Perer, A., Zandifar, A., Visweswaran, S., Eslami, M., Batmanghelich, K.: Medsyn: Text-guided anatomy-aware synthesis of high-fidelity 3d ct images. IEEE Transactions on Medical Imaging  (2024). \doi{10.1109/TMI.2024.3415032}

\bibitem{zhang2023adding}
Zhang, L., Rao, A., Agrawala, M.: Adding conditional control to text-to-image diffusion models (2023)

\bibitem{zhao2025maisiv2accelerated3dhighresolution}
Zhao, C., Guo, P., Yang, D., Tang, Y., He, Y., Simon, B., Belue, M., Harmon, S., Turkbey, B., Xu, D.: Maisi-v2: Accelerated 3d high-resolution medical image synthesis with rectified flow and region-specific contrastive loss (2025), \url{https://arxiv.org/abs/2508.05772}

\bibitem{opensora}
Zheng, Z., Peng, X., Yang, T., Shen, C., Li, S., Liu, H., Zhou, Y., Li, T., You, Y.: Open-sora: Democratizing efficient video production for all. arXiv preprint arXiv:2412.20404  (2024)

\bibitem{zhuangGen}
Zhuang, Y., Hou, B., Mathai, T.S., Mukherjee, P., Kim, B., Summers, R.M.: Semantic image synthesis for abdominal ct. In: Deep Generative Models: Third MICCAI Workshop, DGM4MICCAI 2023, Held in Conjunction with MICCAI 2023, Vancouver, BC, Canada, October 8, 2023, Proceedings. p. 214–224. Springer-Verlag, Berlin, Heidelberg (2023). \doi{10.1007/978-3-031-53767-7_21}, \url{https://doi.org/10.1007/978-3-031-53767-7_21}

\end{thebibliography}

\clearpage
\section{Appendix}

\begin{center}
{\Large \textbf{CONTENTS}}\\[1.5em]
\end{center}

\begingroup
\ttfamily
7.1\quad More Implementation Details \dotfill \\
\hspace*{2.5em}7.1.1\quad Training OSP VAE \dotfill \hyperref[supp:osp]{7.1.1}\\
\hspace*{2.5em}7.1.2\quad Textual Conditions \dotfill \hyperref[supp:text]{7.1.2}\\
\hspace*{2.5em}7.1.3\quad Mask Conditions \dotfill \hyperref[supp:mask]{7.1.3}\\
\hspace*{2.5em}7.1.4\quad Modified DiT \dotfill \hyperref[supp:dit]{7.1.4}\\
\hspace*{2.5em}7.1.5\quad Contrast Windows \dotfill \hyperref[supp:contrast]{7.1.5}\\

7.2\quad Paraphrased Compound Text \dotfill  \hyperref[supp:paraphrase]{7.2}\\

7.3\quad Tokenizer Reconstruction Analysis \dotfill  \hyperref[supp:tokenizer]{7.3} \\

7.4\quad Additional Results \dotfill \\
\hspace*{2.5em}7.4.1\quad Comparing Full-mask Generative Models \dotfill \hyperref[supp:full_mask]{7.4.1}\\
\hspace*{2.5em}7.4.2\quad Radiologist Evaluation \dotfill \hyperref[supp:radiology]{7.4.2}\\
\hspace*{2.5em}7.4.3\quad Augmenting Dataset \dotfill \hyperref[supp:augment]{7.4.3}\\
\hspace*{2.5em}7.4.4\quad Zero-Shot for Extra Abnormality \dotfill \hyperref[supp:zeroshot]{7.4.4}\\

7.5\quad Ablation Study \dotfill \\
\hspace*{2.5em}7.5.1\quad Convolution Layer \dotfill \hyperref[supp:convolution]{7.5.1}\\
\hspace*{2.5em}7.5.2\quad Gated Attention \dotfill \hyperref[supp:gated]{7.5.2}\\





\endgroup

\newpage

\begin{figure}[!]
    \centering
    \includegraphics[width=1.0\linewidth]{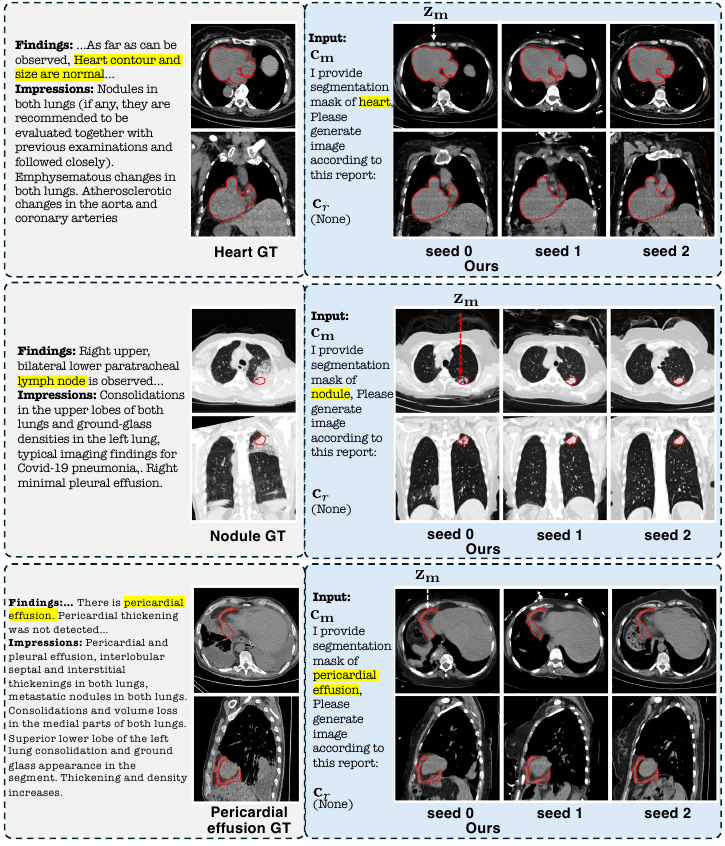}
    \caption{
    \textbf{Results using segmentation masks only.} For each mask, we show ground truth image with contours, and our generated results with three random seeds.
    Our generated image closely follow the given mask, showing both \textbf{controllability} and \textbf{diversity}.
    It allows for spatial detail completion by showing exact heart size (potential cardiomegaly in first row), nodule size (potentially neglected in second row), and pericardial effusion location (third row).
    }
    \label{fig:res_seg_seeds}
\end{figure}

\begin{figure}[!]
    \centering
    \includegraphics[width=1.0\linewidth]{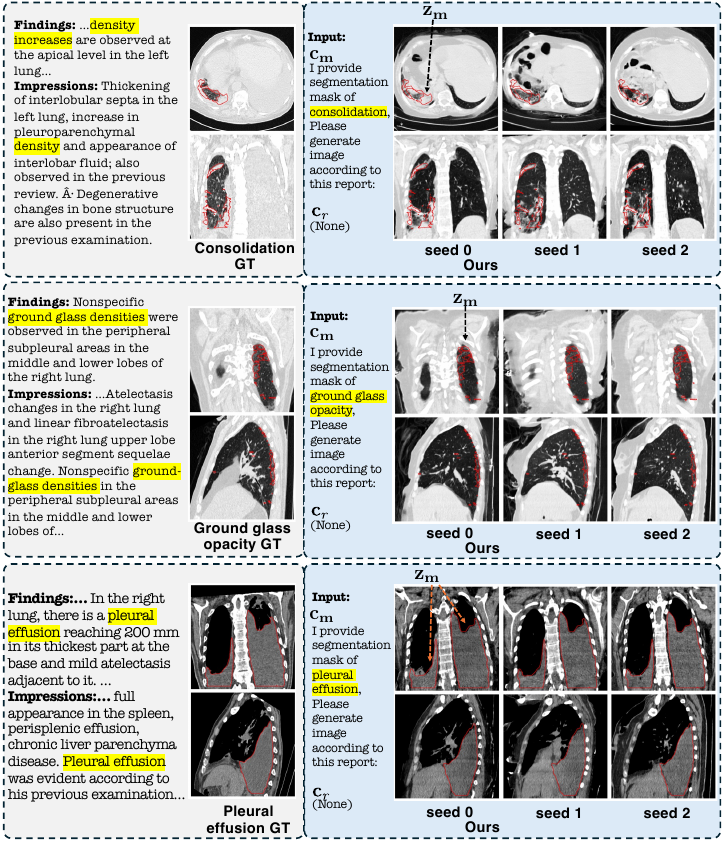}
    \caption{
    \textbf{Results using segmentation masks only.} For each mask, we show ground truth image with contours, and our generated results with three random seeds.
    Our generated image closely follow the given mask, showing both \textbf{controllability} and \textbf{diversity}.
    It allows for spatial detail completion by showing exact consolidation location (first row), ground glass opacity (second row), and pleural effusion location (third row).
    }
    \label{fig:res_abn_seeds}
\end{figure}

\begin{figure}[]
    \centering
    \includegraphics[width=1.0\linewidth]{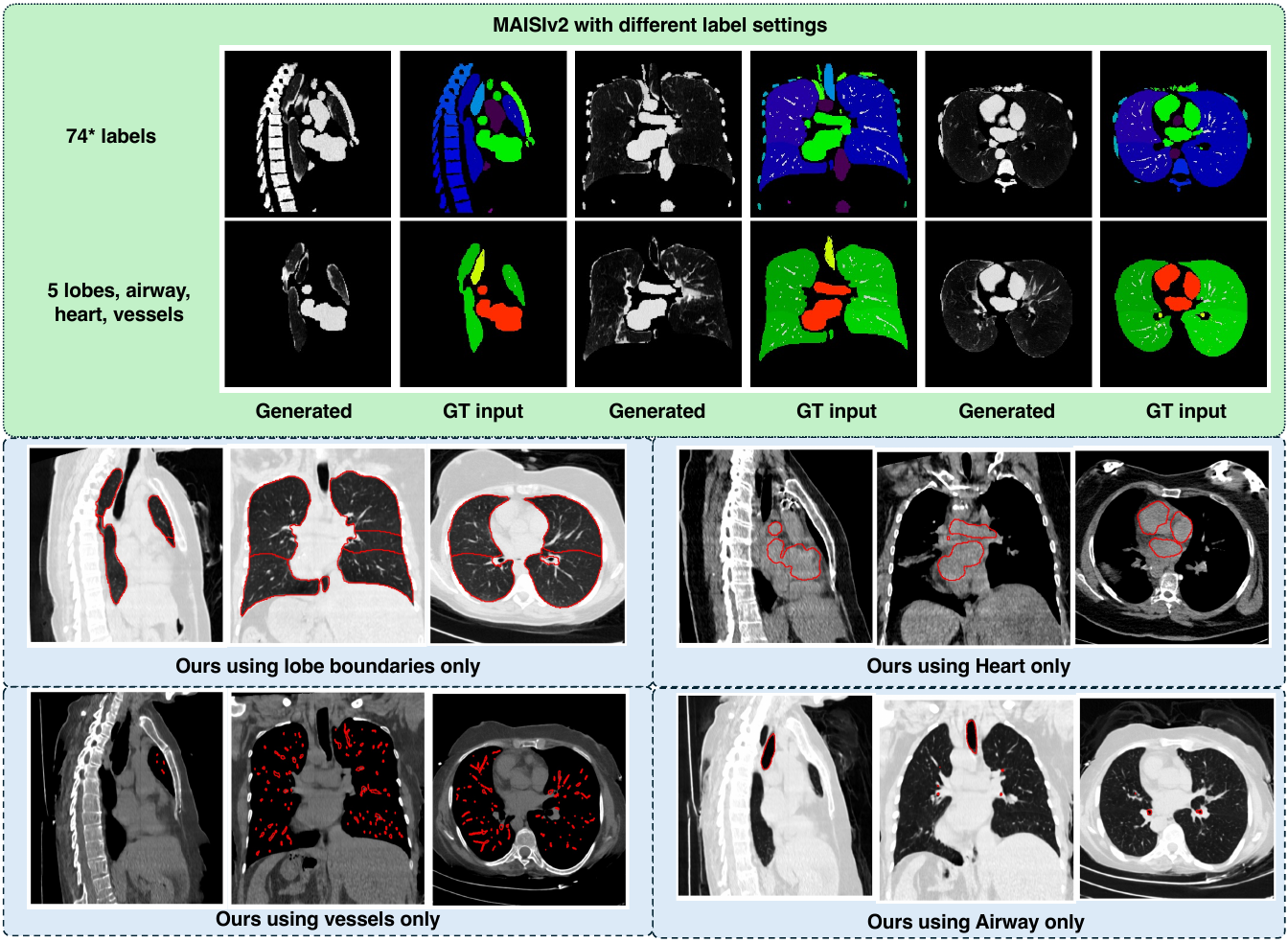}
    \caption{
    \textbf{Results comparing MAISIv2.}
    We show MAISIv2 with two input settings: \emph{1. * the overlap of MAISI labels and TotalSegmentator labels, namely 74 labels, } and \emph{2. using five lobes, airway, heart, vessels (anatomies our model accepts).}
    The results show MAISIv2 fails to generate tissues outside given masks (\eg, soft tissues within skin), leading to its restrictions. 
    Moreover, \textbf{it depends on the \emph{skin mask} to generate soft tissues of human body}, as in this example TotalSegmentator include no skin masks, therefore limited performance.
    Ours however, with single lobe boundaries or single anatomy mask, can successfully synthesize the images.
    This proves the \textbf{controllability} and \textbf{flexibility} of our method.
    }
    \label{fig:maisi}
\end{figure}

\subsection{More Implementation Details}
\textbf{Training OSP VAE.}
\label{supp:osp}
OSP~\cite{opensora, opensora2, lin2024opensoraplanopensourcelarge} was originally designed for video tokenization and decoding. 
We adopt CasualVideoVAE from opensora plan for its superior performance.
The original VAE decodes the latent with $
\mathbf{x} = \text{Dec}(\mathbf{z}) \in 
\mathbb{R}^{(D \times H \times W) },
$
where $D\times H\times W = 449 \times 448 \times 448$, causing some trouble during the sliding window decoding as blurry patch introduced by the extra depth output.
We thus modified the decoder by removing the extra depth concatenation, keeping the number of parameters, while suitable for sliding window inference. 
In conclusion, our encoder works as $\mathbf{z} = \text{Enc}(\mathbf{x}) \in 
\mathbb{R}^{(n_d \times n_h \times n_w) \times n_c}$, where ${(n_d \times n_h \times n_w) \times n_c} = (112\times 56 \times 56)\times 4$; our decoder works as $\mathbf{x} = \text{Dec}(\mathbf{z}) \in 
\mathbb{R}^{(D \times H \times W) }$, where $(D\times H \times W) = (448 \times 448 \times 448)$.

We apply the original training method as adopting reconstruction loss, LPIPS loss, KL-divergence loss, GAN loss, on randomly sampled patch size of $96\times 96 \times 96$, with the training set of CT-Rate dataset~\cite{Hamamci2024GeneralistFM}.
We use Adam optimizer with learning rate $10^{-5}$, and run the training for $2$ epochs.
We analyze the results of inference on both images and masks in Supp.~\ref{supp:tokenizer}.

\noindent \textbf{Textual Conditions.}
\label{supp:text}
Empirically we find all abnormality-related description are tokenized within token length of $24$, thus apply $32$ as the token length for our pre-prompt ($\mathbf{c}_\mathbf{m}$).
For radiology reports, we use a \texttt{<SEP>} token concatenated between impressions and findings. 
We empirically observe that $\sim 97\%$ of reports are within token length $512$, thus adopt this number as our max token length.
Overall, we achieve token length of $544$ by concatenating both (or $32$ when report is missing).

During training, we randomly choose from pre-defined template in Supp.\ref{supp:paraphrase} for all 
four anatomies (\emph{lobe, airway, lung vessel, heart}) and five abnormalities (\emph{consolidations, ground glass opacity, lung nodules, pericardial effusion, pleural effusion}).
However, during testing, we only apply the first template of both (a) and (b) in Fig.~\ref{fig:format}, to render stable results and evaluations.

\noindent \textbf{Mask Conditions.}
\label{supp:mask}
Lung lobe mask is the only mask that exhibit multi-classes, namely left upper, left lower, right upper, right middle, right lower.
To convert it the same as other binary masks, we extract the boundaries followed by dilation with $3\times 3\times 3$, to convert semantic masks to boundary masks. 
We provide analysis below (Supp.~\ref{supp:tokenizer}).
Since our encoder is applied on both images and masks, they should share similar intensity distributions. 
Empirically, we witness using a mapping function into [background, foreground] = $[-0.5, 0.5]$ prevents outliers and enhances distribution similarity.
We thus apply our encoder trained with Supp.~\ref{supp:imple} on all masks to achieve latents $\mathbf{z}$.

Empirically we found all latents exhibit large bias and variance, due to the KL-divergence loss during training. 
Thus to alleviate our LDM burden, we z-norm all latents with respect to each channel, and perform the reversed z-norm during decoding. 

\noindent \textbf{Modified DiT.}
\label{supp:dit}
AdaLN involves the three groups of ~\emph{scale$_1$, shift, scale$_2$} produced by time-embeddings denoted in Fig.~\ref{fig:overview}. 
We follow the original DiT~\cite{Peebles2022DiT} operation here, as $h^j = (\text{scale}_1 \times h^j + \text{shift})\times \text{scale}_2$.
Such operations are adopted three times, surrounding the \emph{MHSA, MHCA, FFN}.

Our convolution layer (Fig.~\ref{fig:overview}, Eq.~\ref{eq:model} last line) involves a $3\times3\times3$ kernel , and then another $1\times1\times1$ kernel. 
Both kernels produce outputs of the same dimensions as the inputs.
We provide ablation studies blow (Supp.~\ref{supp:convolution})

We train our modified DiT model with an AdamW optimizer for $2$ epochs, taking around $24$ hours.

\noindent \textbf{Contrast Windows.}
\label{supp:contrast}
We use different intensity range clipping adopted by radiologists to increase contrast for Lung, Vessel, Bone and Soft Tissue.
The details are [Window Level, Window Width]= Lung: $[-600, 1500]$, Vessel: $[100, 700]$, Soft Tissue: $[50, 350]$, Bone: $[300, 2000]$.
During computing FID scores, we clip the images within respective contrast windows, and pass them into a radimagenet-ResNet50 model~\cite{radimagenet} for computing each axis.

\label{supp:imple}
\subsection{Paraphrased Compound Text}
\label{supp:paraphrase}

\begin{figure}
    \centering
    \includegraphics[width=1.0\linewidth]{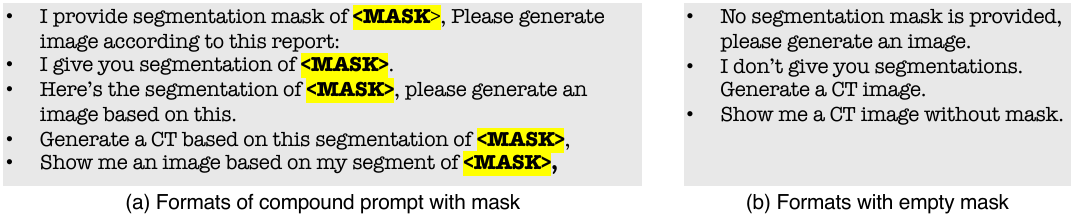}
    \caption{\textbf{Compound prompt format.} (a): the formats with mask, where we use five variances. (b): the formats with no mask provided, where we use three variances. }
    \label{fig:format}
\end{figure}

We provide the paraphrased compound texts shown in Fig.~\ref{fig:format} for all masks during training.
During test time, we only use the first sentence of both (a) and (b). 

\subsection{Tokenizer Reconstruction Analysis}
\label{supp:tokenizer}
\begin{figure}[t!]
\centering
\begin{minipage}{0.45\linewidth}
    \centering
    \includegraphics[width=\linewidth]{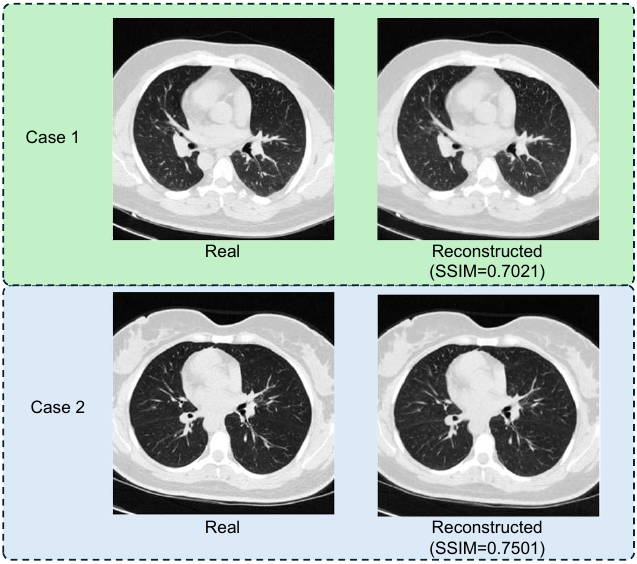}
\caption{
\textbf{Reconstruction performance of images.}
We show 2D slices and 3D SSIM score.}
\label{fig:recon}
\end{minipage}
\hfill
\begin{minipage}{0.48\linewidth}
\centering
\captionof{table}{\textbf{Reconstruction performance of segmentations.}
Dice and Jaccard are reported in \%, HD95 and ASD in voxel (lower is better).
*\emph{We encode lobe with boundaries, thus unable to compare, see Supp.~\ref{supp:imple}}.
}
\label{tab:seg_recon_gt}
\begin{adjustbox}{max width=\linewidth}
\begin{tabular}{lcccc}
\toprule
\textbf{Object}
& Dice $\uparrow$
& Jaccard $\uparrow$
& HD95 $\downarrow$
& ASD $\downarrow$ \\
\midrule
Lobe*                  & - & - & - & - \\
Airway                & 84.89 & 74.22 & 2.26  & 0.76 \\
Vessel                & 62.31 & 45.35 & 2.77  & 0.69 \\
Heart                 & 97.50 & 95.12 & 1.53  & 0.58 \\
\midrule
Ground Glass Opacity  & 88.44 & 79.29 & 1.73  & 0.76 \\
Consolidation         & 64.64 & 51.87 & 10.85 & 0.75 \\
Pleural Effusion      & 80.77 & 71.33 & 2.08  & 0.98 \\
Pericardial Effusion  & 82.45 & 71.64 & 1.99  & 0.78 \\
Nodule                & 63.49 & 49.32 & 3.57  & 0.78 \\
\bottomrule
\end{tabular}
\end{adjustbox}
\end{minipage}
\end{figure}

We analyze our pretrained tokenizer (VAE) based on both images and segmentations. 
We experiment on the test set of CT-Rate, and show reconstruction results on both images and segmentations.
Fig.~\ref{fig:recon} show the axial slices of reconstructed 3D volumes, as well as high SSIM scores on 3D volumes. 
Table~\ref{tab:seg_recon_gt} show the reconstructed segmentation results. We witness high similarity facing large anatomy and abnormalities. 
However on those small areas (\eg, nodule, vessel), the reconstructed results are a little bit off in Dice. 
Their HD95 and ASD still work well, indicating a decent segmentation mask reconstruction in 3D point cloud space.  



\begin{figure}[!]
    \centering
    \includegraphics[width=1\linewidth]{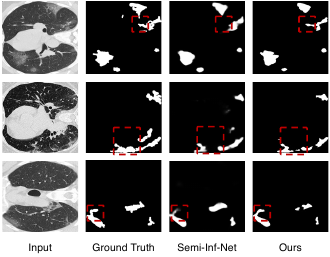}
    \caption{\textbf{Comparison of segmentation results.}
    We compare Semi-Inf-Net results and our vanilla UNet trained with a combination of real data and synthesized data (466 images in total).
    Our method clearly segment delicate details in abnormalities, shown in \textcolor{red}{bounding boxes}. 
    }
    \label{fig:semiseg}
\end{figure}

\begin{table*}[t]
\centering
\caption{\textbf{Segmentation performance of Ours.}
Dice and Jaccard are reported in \%, HD95 and ASD in mm (lower is better).}
\label{tab:seg_comparison}
\begin{adjustbox}{max width=\textwidth}
\begin{tabular}{lcccc|cccc|cccc}
\toprule
 & \multicolumn{4}{c}{\textbf{Generated vs GT}}  & \multicolumn{4}{c}{\textbf{Reconstructed vs GT}}   & \multicolumn{4}{c}{\textbf{Generated vs Reconstructed}} \\
\cmidrule{2-5} \cmidrule{6-9}  \cmidrule{10-13} 
\textbf{Object}
& Dice $\uparrow$ & Jaccard $\uparrow$ & HD95 $\downarrow$ & ASD $\downarrow$
& Dice $\uparrow$ & Jaccard $\uparrow$ & HD95 $\downarrow$ & ASD $\downarrow$
& Dice $\uparrow$ & Jaccard $\uparrow$ & HD95 $\downarrow$ & ASD $\downarrow$ \\
\midrule
Lobe   
& \textbf{93.44} & \textbf{89.16} & \textbf{10.21} & \textbf{3.19} & -- & -- & -- & -- & -- & -- & -- & --  \\
Airway 
& \textbf{69.07} & \textbf{53.73} & \textbf{41.22} & \textbf{4.97} & 84.89 & 74.22 & 2.26 & 0.76 & 68.50 & 52.93 & 41.96 & 4.63 \\
Vessel 
& \textbf{54.82} & \textbf{37.76} & \textbf{44.55} & \textbf{9.11} & 62.31 & 45.35 & 2.77 & 0.69 & 53.80 & 36.80 & 3.16 & 2.39 \\
Heart  
& \textbf{86.40} & \textbf{77.28} & \textbf{16.12} & \textbf{4.88} & 97.50 & 95.12 & 1.53 & 0.58 & 86.81 & 77.48 & 17.55 & 5.28 \\
\midrule
Ground Glass Opacity 
& \textbf{74.15} & \textbf{58.92} & 
\textbf{13.15} & \textbf{4.10} & 88.44 & 79.29 & 1.73 & 0.76 & 64.60 & 47.71 & 17.72 & 3.23 \\
Consolidation 
& \textbf{40.08} & \textbf{37.23} & 
\textbf{86.23} & \textbf{66.83} & 64.64 & 51.87 & 10.85 & 0.75 & 45.44 & 29.40 & 32.29 & 8.55 \\
Pleural Effusion 
& \textbf{82.79} & \textbf{80.32} & \textbf{40.06} & \textbf{21.34} & 80.77 & 71.33 & 2.08 & 0.98 & 78.73 & 64.92 & 14.17 & 5.11 \\
Pericardial Effusion 
& \textbf{39.01} & \textbf{24.23} & \textbf{27.92} & \textbf{13.58} & 82.45 & 71.64 & 1.99 & 0.78 & 70.99 & 55.03 & 24.41 & 4.67 \\
Nodule 
& \textbf{41.77} & \textbf{40.54} & \textbf{94.04} & \textbf{60.14}  & 63.49 & 49.32 & 3.57 & 0.78   & 42.03 & 40.81 & 95.17 & 60.88 \\
\bottomrule
\end{tabular}
\end{adjustbox}
\end{table*}

\subsection{Additional Results}
\label{supp:full_mask}
\noindent \textbf{Comparing Full-mask Generative Models}
We compare against full-mask based generative models, namely MAISIv2~\cite{zhao2025maisiv2accelerated3dhighresolution, maisiv1}, as 3D-MedDiffusion~\cite{Wang20243DMA} does not provide checkpoints regarding mask-based generation, neither do DGM~\cite{xing2024deep} and Trace~\cite{TRACE}.
Results in Fig.~\ref{fig:maisi} show MAISIv2 fails to generate a plausible CT image even with $74$ classes of label, let alone with only lobes, heart, vessel, and airway.
Ours however, successfully synthesizes the CT with only single mask.
This proves our controllability and flexibility.

\noindent \textbf{Radiologist Evaluation.}
\label{supp:radiology}
We include the details of table~\ref{tab:human_eval} in table~\ref{tab:human_eval_extended}. 
As discussed in main text, both radiologists agree on a high consistency between images and their corresponding report and masks. 
As we investigate the details per abnormality, it is obvious that synthetic images suffer from \emph{consolidation} specifically.
After a post discussion, we understand that \emph{consolidations} are always surrounded by \emph{GGO} here, thus making these two hard to discriminate. 
However, among the other three abnormalities, ours even surpass the real images, in that we acquire more \emph{Match} and \emph{Partial Match} counts.
This detailed discussion demonstrates the controllability of our model.

\begin{table}[t!]
\centering
\small
\caption{\textbf{Expert evaluation} of images matching report and mask across pathologies.}
\label{tab:human_eval_extended}

\begin{adjustbox}{max width=\linewidth}
\begin{tabular}{ll lccc}
\toprule
Radiologist & Data & Pathology & Match & \makecell{Partial\\Match} & \makecell{Not\\Match} \\
\midrule

\multirow{10}{*}{\makecell{Radiologist 1\\(4 yrs)}} 
& \multirow{5}{*}{\rotatebox{90}{Real}}
& Consolidation & 10 & 2 & 3 \\ 
& & GGO & 10 & 1 & 4 \\ 
& & Nodule & 8 & 0 & 7 \\ 
& & Pericardial Effusion & 12 & 1 & 2 \\ 
& & Pleural Effusion & 9 & 6 & 0 \\ \cline{2-6}

& \multirow{5}{*}{\rotatebox{90}{Synthetic}}
& Consolidation & 3 & 5 & 7 \\
& & GGO & 6 & 5 & 4 \\ 
& & Nodule & 8 & 3 & 4 \\ 
& & Pericardial Effusion & 11 & 2 & 2 \\ 
& & Pleural Effusion & 12 & 3 & 0 \\

\midrule

\multirow{10}{*}{\makecell{Radiologist 2\\(17 yrs)}}
& \multirow{5}{*}{\rotatebox{90}{Real}}
& Consolidation & 4 & 9 & 2 \\ 
& & GGO & 9 & 5 & 1 \\ 
& & Nodule & 8 & 6 & 1 \\
& & Pericardial Effusion & 10 & 4 & 1 \\ 
& & Pleural Effusion & 14 & 1 & 0 \\ \cline{2-6}

& \multirow{5}{*}{\rotatebox{90}{Synthetic}}
& Consolidation & 0 & 9 & 6 \\
& & GGO & 3 & 11 & 1 \\ 
& & Nodule & 10 & 4 & 1 \\ 
& & Pericardial Effusion & 11 & 3 & 1 \\ 
& & Pleural Effusion & 12 & 3 & 0 \\

\bottomrule
\end{tabular}
\end{adjustbox}

\end{table}

\noindent \textbf{Augmenting Dataset.}
\label{supp:augment}
We provide the segmentation visualization results in Fig.~\ref{fig:semiseg}.
This is an extension of our main text Table~\ref{tab:seg_recon_gt}.
Our method includes a combination of both real data and synthesized data, resulting in $466$ 2D images, on which the trained vanilla UNet even beats current SOTA methods. 
The red bounding box in Fig.~\ref{fig:semiseg} shows our method generates better continuity among abnormality masks.
This results from the synthesized data with high quality and fidelity, exposing models to various abnormality features, leading to a better understanding with the combination of real data.

\begin{figure}[t!]
    \centering
    \includegraphics[]{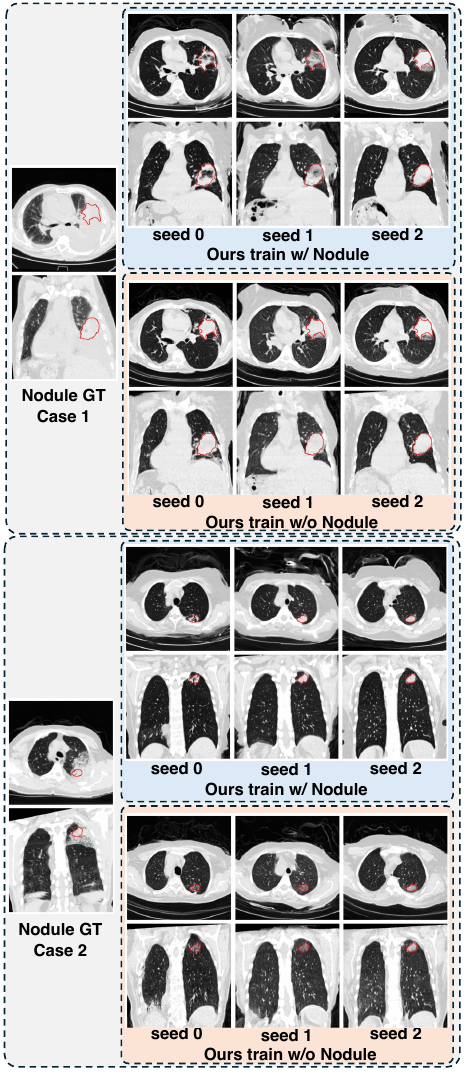}
    \caption{
    \textbf{Results of training without nodule labels.}
    We train another model without nodule labels.
    For each case, we show original results on the top, and \textbf{training without nodule on the bottom}.
    In case 1 where masks are huge, nodules can still be synthesized, even surpassing the original quality. 
    However in case 2 where the nodules are smaller, the results are less denser, leading to poor quality.
    }
    \label{fig:ood}
\end{figure}

\noindent \textbf{Zero-Shot Extra abnormality.}
\label{supp:zeroshot}
We train a model without including nodule labels.
That leaves to lobe, vessel, airway, heart; consolidation, ground glass opacity, pericardial effusion, pleural effusion; in total four anatomies and four abnormalities.
The results are shown in Fig.~\ref{fig:ood}.
We show two cases, within which we show \emph{original results} on the top, and \emph{training without nodule} on the bottom.
Both cases are inferred without radiology to reduce impact; however in $\mathbf{c}_\mathbf{m}$ we reveal the identity of nodule.
In case 1, \emph{training without nodule} even produces nodules of better quality, with more uniform tissue characteristic. 
However in case 2 where the nodules are smaller, the synthesized nodules show worse quality.
This experiments show:
1) even training without nodule can still somehow generate corresponding abnormality; we attribute it to our text encoder absorbing knowledge from radiology reports;
2) however when it comes to more realistic scenario (\eg, nodules are often tiny in lung CTs), results show some hallucination; we conclude that our model still show some good results in a zero-shot setting, but the performance is limited and therefore corresponding masks training are preferred.

\subsection{Ablation Study}
\begin{figure}[t!]
    \centering
    \includegraphics[width=1.0\linewidth]{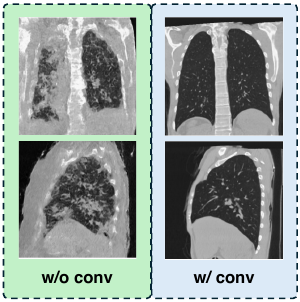}
    \caption{
    \textbf{Ablation of \textit{conv}.} 
    Results \textit{w/o} conv layer show obvious quality drawback, while with conv layer it performs good.
    }
    \label{fig:res_abla_conv}
\end{figure}

\noindent \textbf{Convolution Layer.}
\label{supp:convolution}
We provide qualitative results of with convolution layer and without convolution layer in Fig.~\ref{fig:res_abla_conv}.
Clear artifact exists in the results without convolution layers, as only partial bones and partial lung structures are preserved.
However, when we add the convolution layers, results perform better as more details exhibit high resolution quality, \eg, vessels, ribs.
We attribute this to our convolution layer, where stronger inter-patch communication alleviates the drawback of information loss in volumetric domains, leading to better performance.
This ablation study proves the necessity of adopting our convolution layer, as generating more realistic and fine-grained volumetric lung CT images..

\begin{figure}[!]
    \centering
    \includegraphics[width=1.0\linewidth]{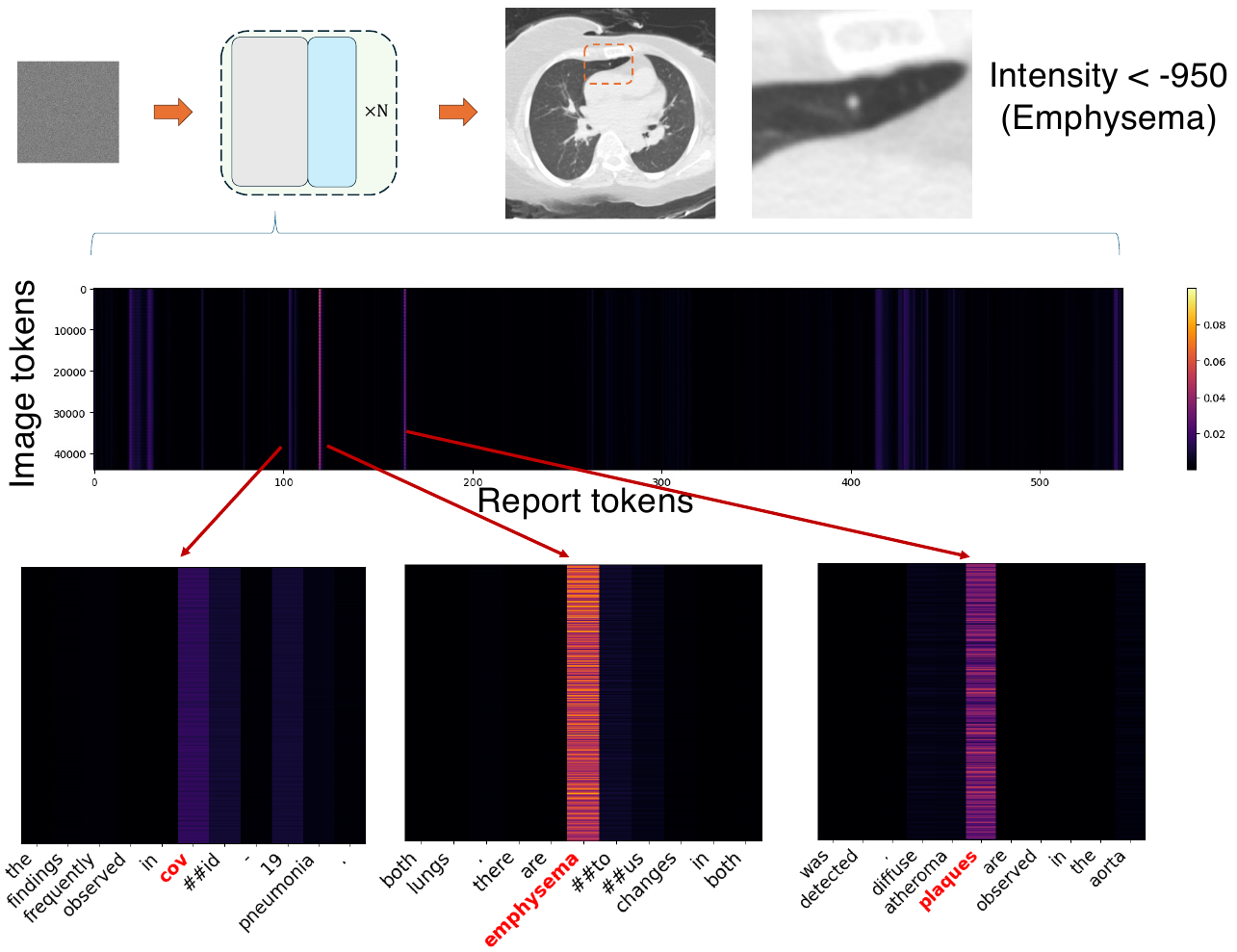}
    \caption{
    \textbf{Cross Attention between report tokens (x-axis) and image tokens (y-axis) of model \emph{with gated attention}.} 
    We average the cross attention of all attention heads and layers and show examples of highly correlated image tokens and text tokens. 
    Our model emphasizes pathological terms, \textit{e.g.,} `Emphysema', `Covid-19', `plaques'.
    Thus the generated image (top) show clear characteristic of emphysema.
    }
    \label{fig:res_crossattn}
\end{figure}

\begin{figure}
    \centering
    \includegraphics[width=1\linewidth]{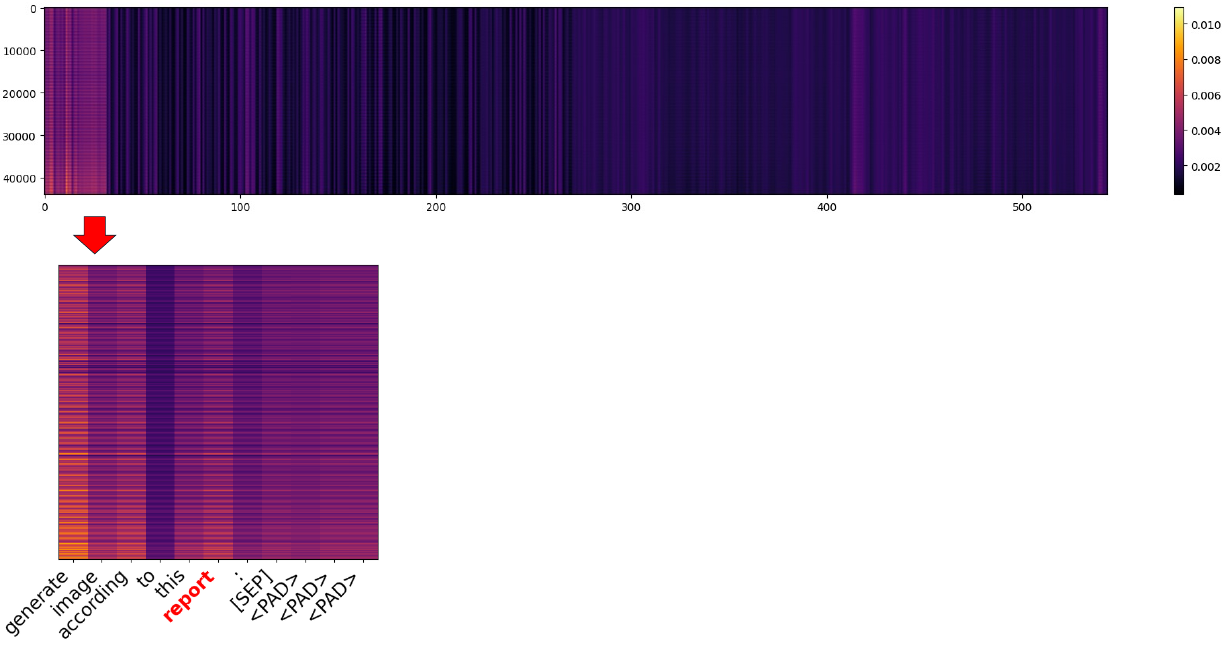}
    \caption{
    \textbf{Cross Attention between report tokens (x-axis) and image tokens (y-axis) of model \emph{without gated attention}.}
    Similarly, we average all attention heads and layers, yet this time the model 1) pays more attention to the beginning of the text, (\ie, the same problem raised in~\cite{qiu2025gated}); 2) shows more uniformed attention map in later tokens, instead of focusing on abnormality terms.
    }
    \label{fig:res_crossattn_base}
\end{figure}

\noindent \textbf{Gated Attention.}
\label{supp:gated}
We train another model with the same settings, except we remove all gated attention modules from both MHSA and MHCA, and compare it against our original model.
We provide the same instruction to both models (including $\textbf{c}_\textbf{m}$ and $\textbf{c}_r$) and show the averaged cross attention value of all heads in all layers. 
Fig.~\ref{fig:res_crossattn} shows results with gated attention, while Fig.~\ref{fig:res_crossattn_base} shows results without gated attention.
The result with attention shows our model successfully focuses on the abnormality related terms (\ie covid-19 pneumonia, emphysema, atheroma plaques), and outputs corresponding emphysema lung tissues, proving the efficacy of receiving instructions from text and generating corresponding features.
However, Fig.~\ref{fig:res_crossattn_base} demonstrates that if without gated attention, 1) the model will pay more attention to leading tokens, as also illustrated in Gated Attention~\cite{qiu2025gated}; 2) the model shows a uniformed distribution in the radiology report tokens, resulting in worse ability of capturing abnormality terms.
This ablation study emphasizes that our gated attention incorporation boosts the understanding of radiology reports, and corresponding CT generation.


\end{document}